\documentclass[acmtog,screen,nonacm,balance=false]{acmart}

\usepackage{booktabs} 
\usepackage[export]{adjustbox}
\usepackage{caption}
\usepackage{subcaption}

\setcitestyle{nosort,square} 

\title[PSDR-Room: Single Photo to Scene using Differentiable Rendering]{%
    \sysName: Single Photo to Scene using Differentiable Rendering
}

\author{Kai Yan}
\affiliation{%
    \institution{University of California, Irvine}%
    \country{USA}
}	
\affiliation{%
    \institution{Adobe Research}%
    \country{USA}
}	

\author{Fujun Luan}
\author{Miloš Hašan}
\author{Thibault Groueix}
\affiliation{%
    \institution{Adobe Research}%
    \country{USA}
}
\author{Valentin Deschaintre}
\affiliation{%
    \institution{Adobe Research}%
    \country{UK}
}
\author{Shuang Zhao}
\affiliation{%
    \institution{University of California, Irvine}%
    \country{USA}
}

\renewcommand\footnotetextcopyrightpermission[1]{} 

\usepackage{mathtools,stmaryrd,nicefrac,bbm}
\usepackage[many]{tcolorbox}
\usepackage{ifthen,comment,xspace}
\usepackage{overpic}
\usepackage[outline]{contour}
\contourlength{0.075em}

\usepackage[linesnumbered,ruled,vlined]{algorithm2e} 

\SetAlFnt{\small}
\SetAlCapFnt{\small}
\SetAlCapNameFnt{\small}
\SetAlCapHSkip{0pt}
\let\oldnl\nl
\newcommand{\nonl}{\renewcommand{\nl}{\let\nl\oldnl}}

\usepackage{tikz}
\usetikzlibrary{calc}
\usepackage[normalem]{ulem}

\usepackage{enumitem}
\setlist[itemize]{parsep=0pt,partopsep=0pt,leftmargin=*,itemsep=5pt}
\setlist[enumerate]{parsep=0pt,partopsep=0pt,leftmargin=*,itemsep=5pt}

\definecolor{orange_cubic}{rgb}{.9765, .5887, .3569}
\definecolor{purple_cubic}{rgb}{.4706, 0, .5216}
\definecolor{green_cubic}{rgb}{.28603, .81178, .5008}

\definecolor{grayLL}{rgb}{.98, .98, .98}
\definecolor{grayL}{rgb}{.9, .9, .9}
\definecolor{purpleL}{rgb}{.9735, .95, .9761}
\definecolor{purpleD}{rgb}{.8941, .8, .9043}
\definecolor{greenL}{rgb}{.9643, .9906, .9750}
\definecolor{greenD}{rgb}{.7145, .9249, .7999}
\definecolor{greenDD}{rgb}{.3145, .6249, .3999}
\definecolor{orangeLL}{rgb}{0.9991, 0.9846, 0.9759}
\definecolor{orangeL}{rgb}{.9982, .9692, .9518}
\definecolor{orangeD}{rgb}{.9929, .8766, .8071}

\definecolor{redL}{rgb}{1.0, 0.95, 0.95}
\definecolor{redD}{rgb}{1.0, 0.8, 0.8}
\definecolor{redDD}{rgb}{1.0, 0.4, 0.4}
\definecolor{yellowL}{rgb}{1.0, 1.0, 0.95}
\definecolor{yellowD}{rgb}{0.95, 0.95, 0.6}
\definecolor{yellowDD}{rgb}{0.8, 0.8, 0.2}
\definecolor{blueLL}{rgb}{0.98, 0.98, 1.0}
\definecolor{blueL}{rgb}{0.95, 0.95, 1.0}
\definecolor{blueD}{rgb}{0.8, 0.8, 1.0}
\definecolor{blueDD}{rgb}{0.6, 0.6, 1.0}

\definecolor{uciBlue}{RGB}{0,100,164}
\definecolor{uciBlueL}{RGB}{127.5, 177.5, 209.5}
\definecolor{uciOrange}{RGB}{247,141,45}
\definecolor{uciOrangeL}{RGB}{251,198,150}

\newtcolorbox{myBox}{%
	colback=grayLL,colframe=grayL,top=1mm,bottom=1mm,left=1mm,right=1mm%
}
\newtcolorbox{myTitledBox}[2][]{%
	colback=grayLL,colframe=grayL,top=1mm,bottom=1mm,left=1mm,right=1mm,enlarge top by=0.5em,title={#2},fonttitle=\bfseries\small\color{gray},#1%
}

\tcolorboxenvironment{myitemizeP}{blanker,before skip=6pt,after skip=6pt,borderline west={3mm}{0pt}{purpleD}}

\tcolorboxenvironment{myitemizeG}{blanker,before skip=6pt,after skip=6pt,borderline west={3mm}{0pt}{greenD}}

\tcolorboxenvironment{myitemizeO}{blanker,before skip=6pt,after skip=6pt,borderline west={3mm}{0pt}{orangeD}}
\newtcbtheorem{thm}{Theorem}%
{colback=white,colframe=orangeD,top=1mm,bottom=1mm,left=1mm,right=1mm,enlarge top by=1mm,fonttitle=\bfseries\color{black}}{thm}
\newcommand{\mymathbox}[3]{%
    \tcboxmath[top=0mm,bottom=0mm,left=0mm,right=0mm,fonttitle=\bfseries\scriptsize\color{gray},colbacktitle=white,enhanced,attach boxed title to top center={yshift=-1mm},boxed title style={top=0mm,bottom=0mm,left=0mm,right=0mm},colframe=#1,colback=white,title=#2]{#3}
}

\newcommand{\defeq}{\vcentcolon=}

\makeatletter
\newcommand{\dotr}[1]{%
	\mathpalette\@dotr{#1}%
}
\newcommand*{\@dotr}[2]{%
	\sbox0{$\m@th#1#2$}%
	\usebox{0}%
	\raisebox{\dimexpr\ht0-\height}{$\m@th#1\@smallbullet#1\bullet$}%
	\kern\scriptspace
}
\newcommand*{\@smallbullet}[2]{%
	\scalebox{.4}{$\m@th#1#2$}%
}
\makeatother

\newcommand{\sysName}{\textsf{PSDR-Room}\xspace}

\newcommand{\bx}{\boldsymbol{x}}

\newcommand{\bp}{\boldsymbol{p}}

\newcommand{\bI}{\boldsymbol{I}}

\newcommand{\Real}{\mathbb{R}}

\newcommand{\calB}{\mathcal{B}}
\newcommand{\calL}{\mathcal{L}}
\newcommand{\calM}{\mathcal{M}}

\newcommand{\ttx}{\mathtt{X}}

\newcommand{\D}{\mathrm{d}}
\newcommand{\pathspace}{\boldsymbol{\Omega}}
\newcommand{\pathspaceU}{\hat{\pathspace}}
\newcommand{\dpathspaceU}{\partial\pathspaceU}
\newcommand{\lightpath}{\bar{\bx}}
\newcommand{\lightpathU}{\bar{\bp}}
\newcommand{\f}{f}
\newcommand{\fU}{\hat{\f}}

\newcommand{\param}{\theta}
\newcommand{\bparam}{\boldsymbol{\theta}}

\newcommand{\mparam}{m_{\bparam}}

\newcommand{\tinterior}{\emph{interior}\xspace}
\newcommand{\tboundary}{\emph{boundary}\xspace}
\newcommand{\vel}{V}
\newcommand{\dmu}{\dot{\mu}}

\newcommand{\diffeqn}[2]{%
    \mymathbox{uciBlueL}{interior}{#1} + \mymathbox{uciOrangeL}{boundary}{#2}%
}

\newcommand{\refdef}[2]{%
    \hyperlink{label:def:#1}{\textcolor{gray}{#2}}%
}

\newlength{\tikzLen}

\newlength{\resLen}

\newcommand{\psdrcuda}{PSDR-CUDA\xspace}
\newcommand{\loss}{\mathcal{L}}

\definecolor{amber}{rgb}{1.0, 0.75, 0.0}

\newcommand{\sz}[1]{}
\newcommand{\vde}[1]{}
\newcommand{\mh}[1]{}
\newcommand{\ky}[1]{}
\newcommand{\fl}[1]{}
\newcommand{\tg}[1]{}
\newcommand{\rev}[1]{{#1}}
\newcommand{\revky}[1]{{#1}}

\begin{document}

\begin{teaserfigure}
    \includegraphics[width=\textwidth]{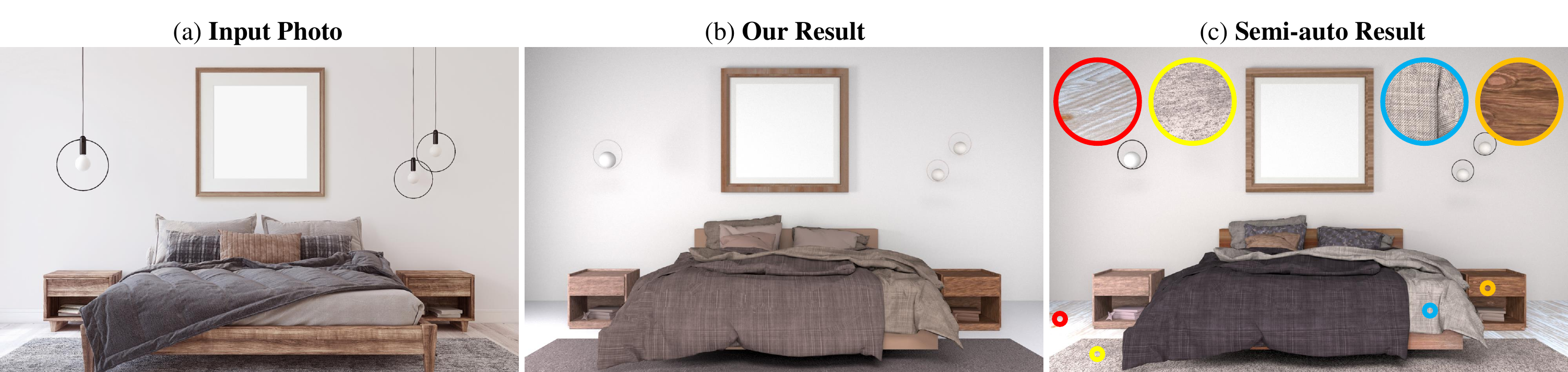}
    \caption{Given an input photo of an indoor scene (a), PSDR-Room automatically retrieves, initializes and optimizes geometry,
    procedural materials and illumination to closely match the scene appearance. The reconstructed result (b) is a valid 3D scene, with editable camera, lighting, geometry, and materials. We provide an option for the user to pick a list of corresponding crop rectangles for appearance matching, resulting in slightly better color/material matching, as shown in our semi-auto result (c). Our method leverages optimization using path-space differentiable rendering, combined with initialization based on depth and segmentation neural models, and object and material search based on CLIP-space proximity.}
     \label{fig:teaser}
 \end{teaserfigure}
 
\begin{abstract}
    A 3D digital scene contains many components: lights, materials and geometries, interacting to reach the desired appearance. Staging such a scene is time-consuming and requires both artistic and technical skills. In this work, we propose \sysName, a system allowing to optimize lighting 
    as well as the pose and materials of individual objects 
    to match a target image of a room scene, with minimal user input.
    To this end, we leverage a recent path-space differentiable rendering approach that provides unbiased gradients of the rendering with respect to geometry, lighting, and procedural materials, allowing us to optimize all of these components using gradient descent to visually match the input photo appearance.
    We use recent single-image scene understanding methods to initialize the optimization and search for appropriate 3D models and materials. We evaluate our method on real photographs of indoor scenes and demonstrate the editability of the resulting scene components.
\end{abstract}

\maketitle
\section{Introduction}
\label{sec:intro}
Progress in computer graphics over recent decades has led to algorithms that turn detailed scene descriptions into highly photorealistic renderings. Such a scene description is composed of many components: lights, materials and geometries (themselves separated into sub-geometries with different material assignments). These pieces all interact to reach the desired appearance. However, the problem of composing such scenes (especially with complex materials and lighting) is a manual undertaking that requires significant technical and artistic skill, creating a need for methods that automatically handle parts of the arduous process of 3D content creation.

Composing a scene from a single photograph with a high level of automation is a long-standing challenge~\cite{Roberts1963}. Several recent work made progress on parts of the problem, focusing either on geometry retrieval and placement~\cite{im2cad, total3D} or materials and lighting~\cite{photoscene} independently. Recently, progress in physically-based differentiable rendering reached a point where full light transport computation can also provide additional \emph{gradients} of the final pixel values with respect to all relevant scene parameters: geometric information, material parameters, and lighting parameters. For the first time, this theoretically opens the exciting possibility of jointly refining the estimates of all scene components through optimization.

In this work, we propose a system, \sysName, to match the appearance of a single photograph of an indoor scene by initializing and optimizing all of these components: lighting, geometry and materials. To do so, we leverage recent path-space differentiable rendering (PSDR) approaches \cite{Zhang:2020:PSDR,Zhang:2021:PSDR,Yan:2022:Guiding}. We also take advantage of the recent progress in scene understanding, using powerful models for image segmentation~\cite{Mask2Former}, depth estimation~\cite{ranftl2020towards, ranftl2021vision}, FOV estimation~\cite{perspectivefields}, and image latent-space encoding~\cite{CLIP}. Furthermore, given their importance in industry, we retrieve and optimize materials represented as procedural node graphs, using a recent differentiable approach~\cite{MATch, li2023diffmatv2}.
With our system, given an input photograph, users can generate a visually matching 3D scene within a few minutes, using geometry and procedural material assets from their preferred library. Once optimized, the scene can easily be edited to change its components, or further modify the parameters of the selected procedural materials. Throughout our pipeline, differentiable rendering is crucial in adjusting initial guesses and reaching a high quality match to the target image.

Our method 
involves several stages as follows.  First, the preprocessing step leverages state-of-the-art single image estimators for camera intrinsics, depth and object segmentation, which combine to estimate a point cloud and divide the scene into its component objects. Based on this initial information, the next stage finds an approximate room shape, which we further refine through physically-based differentiable rendering. The object stage then aims at retrieving, positioning and orienting each object detected by the segmentation. Again, our method first estimates a rough position based on the estimated point cloud and further refines this prediction using differentiable rendering. In the material and lighting stage, we assign a material, possibly based on a complex procedural node graph retrieved from a database, to each object and generate lighting, based on the layout of the scene. Once more, we leverage differentiable rendering to jointly refine all material properties as well as the scene lighting. 

Our system also allows user guidance, by specifying pairs of crop between the target image and a rendering of our optimised scene to enforce material similarity in these areas. As our method is retrieval based, multiple objects of materials can be similar to the target. \revky{In our main setting (automatic), we select the top-1 result, but a user can easily adjust the select object/material among the closest match.}

We evaluate our methods against real indoor scene photographs, obtaining visually close reconstructions. We further compare to recent work, showing that our approach is more accurate in jointly estimating lights and materials. In summary, we propose a method to go from a single photograph to a 3D scene with lighting and procedural materials in a few minutes with a minimal amount of user input. This is the first end-to-end system applying physically-based differentiable rendering to turn an input photo into a valid 3D scene with separate objects, capable of optimizing geometry poses, lighting and complex procedural materials.
\section{Related Works}
\label{sec:related}
\paragraph{Scene-level inverse rendering}
Creating entire 3D scenes from photographs alone is one of the big challenges at the interaction between Computer Graphics and Vision. Previous work leverages deep learning to directly reconstruct meshes~\cite{total3D} or infer per-pixel scene illumination, normal and material parameters~\cite{li2020inverse, zhu2022irisformer}. Other approaches take advantage of the recent progress in differentiable rendering to optimise for texture and materials~\cite{azinovic2019inverse, nimierdavid2021material}. Reconstructing both geometries and materials with a production-level quality is a particularly challenging task. Closer to our approach, different work proposed to instead retrieve geometries in a mesh database and optimise their positions given a target photograph, such as the \textsf{IM2CAD} system~\cite{im2cad}. Other approaches~\cite{total3D,huang2018cooperative} predict geometries from a single photo directly, not utilizing a database; it also predicts poses and camera parameters. As opposed to our approach, these works do not match lighting or materials and do not benefit from the recent progress in differentiable rendering to enable fine-grained adjustment to the generated scene. 

A recent method, \textsf{PhotoScene}~\cite{photoscene}, retrieves and optimizes materials as well as lighting on a pre-existing geometry scene (manually created or predicted by a different method like Total3D~\cite{total3D}). This work is fairly close to our material and lighting stage, and also uses differentiable procedural materials as a texture prior. However, they use simpler approximations for differentiable rendering. Our material and lighting results improve upon theirs, due to having a more powerful full-scene differentiable rendering system, and CLIP-based search for materials. We also find that comparing texture on rectangular crops is more robust than masked VGG losses used by PhotoScene. 

\paragraph{Differentiable rendering}
Specialized differentiable renderers have long existed in computer graphics and vision \cite{gkioulekas2013inverse,gkioulekas2016evaluation,azinovic2019inverse,tsai2019beyond,che2020towards}. Recently, several general-purpose ones like \texttt{redner}~\cite{Li:2018:DMC}, \texttt{Mitsuba~2/3}~\cite{nimier2019mitsuba}, and \texttt{\psdrcuda}~\cite{Zhang:2020:PSDR} have been developed.
A key technical challenge in differentiable rendering is to estimate gradients with respect to object geometry (e.g., positions of mesh vertices). To this end, several approximated methods~\cite{liu2019soft,loubet2019reparameterizing} have been proposed. Unfortunately, inaccuracies introduced by these techniques can lead to degraded result quality, as demonstrated by Luan~et~al.~\shortcite{Luan2021}.
On the contrary, recent techniques specifically sampling or reparameterizing visibility boundaries~\cite{Li:2018:DMC,Zhang:2019:DTRT,Zhang:2020:PSDR,Bangaru:2020:WAS,Zhang:2021:PSDR} provides unbiased gradient estimates capable of producing higher-quality reconstructions.

\paragraph{Procedural material optimization}
Procedural materials are an industry standard for material representation is procedural materials. The manipulation and generation of such materials has been an active field of research in recent years. The most recent work on procedural material are starting to enable procedural graph generation~\cite{hu2022inverse, Guererro22}, however the generated material quality do not yet match the existing databases. Most relevant to our goals are therefore procedural material parameters estimation~\cite{hu2019} and optimization~\cite{MATch, hu2022diff, li2023diffmatv2} methods. Combined with a new material retrieval approach, we leverage this progress in procedural materials optimization to better match the target scene appearance. 

\paragraph{Single image scene understanding}
Recent year saw significant progress in neural models for single image scene understanding. We utilize models for intrinsic camera parameter~\cite{lopez2019deep, zhang2020deepptz, perspectivefields} and depth estimation~\cite{ranftl2020towards, ranftl2021vision}. We benefit from this progress, to better understand the photograph FOV and scene geometry. To automate our method as much as possible, we benefit from recent improvements in instance level segmentation in a single image~\cite{Mask2Former, kirillov2023segment}, allowing us to separate the different geometries both spatially and semantically. Our approach relies on these scene understanding components, but is not directly tied to any specific method, meaning that it would benefit from any future progress in this area. 

\section{Preliminaries}
\label{sec:prelim}
In this section, we introduce several technical tools, recently made available by the research community, which are key enablers of our approach. We give a brief introduction to path-space differentiable rendering \cite{Zhang:2020:PSDR,Zhang:2021:PSDR} used by our \psdrcuda system. We also discuss the inverse procedural material approach MATch and its library of differentiable procedural nodes, DiffMat \cite{MATch}, as well as losses based on Gram matrices of VGG layers \cite{Gatys2015}.

\subsection{Path-Space Differentiable Rendering}
\label{ssec:prelim_psdr}
Physically based rendering is frequently formalized using the path-integral formulation by Veach~\shortcite{veach1997robust}. The intensity $I$ of a final image pixel resulting is an integral over all light paths through that pixel:
\begin{equation}
    \label{eqn:I}
    I = \int_{\pathspace} \f(\lightpath) \,\D\mu(\lightpath),
\end{equation}
where: $ \pathspace$ is the path space comprised of light paths $\lightpath = (\bx_0, \ldots, \bx_N)$ of all lengths $N \geq 1$, connecting the camera to the light source through a number of scene interaction events; $\mu$ is an appropriate measure on the space of paths.
Finally, $\f$ is the \emph{contribution function} defined as the product of terms on each path vertex: \emph{source emission} on the light source vertex; \emph{bidirectional scattering distribution functions} (BSDFs) on vertices corresponding to scene geometry reflection and transmission events; \emph{geometric terms} on the path segments, and a \emph{detector response} term on the camera vertex (typically based on a pixel reconstruction filter such as a box or a Gaussian).


A key challenge is to extend the above formulation to differentiable rendering, where the goal is to compute the derivative of the pixel intensity $I$ with respect to some scene parameter $\param \in \Real$. We could think of $\theta$ as a time-like parameter, where the scene ``evolves'' with it, though this does not need to be the case; there could be many such parameters controlling lighting, materials and geometry.

\iffalse
    In the latter case when scene geometry (i.e., union of object surfaces) $\calM$ evolves with some parameters~$\param \in \Real$, Zhang~et~al.~\shortcite{Zhang:2020:PSDR,Zhang:2021:PSDR} proposed a reparameterization to facilitate the differentiation of Eq.~\eqref{eqn:I} with respect to $\param$. The details of the reparameterization are beyond our scope, but the result is that differentiating Eq.~\eqref{eqn:I} with respect to $\param$ gives
    \begin{equation}
        \label{eqn:dI}
        \frac{\D I}{\D\param} =
        \diffeqn{%
            \int_{\pathspaceU} \frac{\D}{\D\param}\fU(\lightpathU) \,\D\mu(\lightpathU)
        }{%
            \int_{\dpathspaceU} \Delta\fU_K(\lightpathU) \,\vel(\bp_K) \,\D\dmu(\lightpathU)
        }\,,
    \end{equation}
    where $\fU$ is a reparameterized version of $f$. The \tinterior component is obtained by differentiating the integrand contribution function $\fU$.

    The \tboundary integral---which is unique to differentiable rendering---operates over the \emph{boundary path space}~$\dpathspaceU$ that comprises boundary paths $\lightpathU = (\bp_0, \ldots, \bp_N)$.
    These paths are identical to the ordinary ones except for containing exactly one \emph{boundary segment}~$\overline{\bp_K\,\bp_{K + 1}}$ (where $0 \leq K < N$) that lies on a discontinuity of the mutual visibility function.
    Further, $\vel$ captures the scalar ``normal velocity'' of the discontinuity boundary with respect to $\param$. 
\else
    In the latter case when scene geometry (i.e., union of object surfaces) $\calM$ evolves with some parameters~$\param \in \Real$, Zhang~et~al.~\shortcite{Zhang:2020:PSDR,Zhang:2021:PSDR} have demonstrated that the derivative $\nicefrac{\D I}{\D\param}$ of Eq.~\eqref{eqn:I} can be expressed as the sum of an \tinterior and a \tboundary path integrals where the latter is unique to differentiable rendering and capture light transport paths with a segment constrained on a visibility boundary.
\fi

In practice, our renderer \psdrcuda gives correct gradients with respect to parameters $\param$ that cause geometric change, because it appropriately samples the boundary as well as interior terms. Note that this formulation can be easily adapted beyond solving full light transport, and we use it to compute anti-aliased depth and object mask images as well, by modifying the contribution function $\f$ accordingly.

\subsection{Inverse rendering and losses}
\label{ssec:inv_render}
Optimization-based inverse rendering, or analysis by synthesis, infers a set of $\mparam$ scene parameters~$\bparam \in \Real^n$ by minimizing some predetermined rendering loss $\loss$ between the rendered image $\bI$ and reference image $\bI_0$. Solving this optimization 
using stochastic gradient descent methods such as Adam~\cite{kingma2014adam} requires differentiating the rendering loss $\calL$ with respect to the parameters~$\bparam$.
According to the chain rule, the gradient $\nicefrac{\D\calL}{\D\bparam}$ satisfies
\begin{equation}
    \label{eqn:chain}
    \frac{\D\calL}{\D\bparam} = \frac{\partial\calL}{\partial\bI} \,\frac{\D\bI}{\D\bparam},
\end{equation}
where the $\nicefrac{\partial\calL}{\partial\bI}$ on the right-hand side can be obtained using differentiable evaluation of the loss $\calL$, and $\nicefrac{\D\bI}{\D\bparam}$ is computed 
using differentiable rendering (\S\ref{ssec:prelim_psdr}).

\paragraph{Procedural materials}
Substance materials~\cite{SubstanceDes} are an industry standard for defining realistic material textures through procedural node graphs, where nodes generates noises and patterns, and adjust them using image processing filters.
The MATch approach and DiffMat library by Shi et al.~\shortcite{MATch} implemented differentiable versions of many of the filter nodes in the Substance engine, which can be used to optimize their parameters to match a target material appearance. \revky{The DiffMat library was recently improved (DiffMatV2) for faster optimization and generator node optimization~\cite{li2023diffmatv2}.} Similar to PhotoScene~\cite{photoscene}, we use this approach (we leverage the most recent version) to define a manifold of plausible textures, with a much smaller number of optimizable parameters than texels, providing important regularization such that even invisible parts of the scene objects receive valid texture.

\revky{In practice, the presence of procedural material parameters introduces another step into the chain rule above. The differentiable rendering will back-propagate the loss to a texture space gradient, after which the DiffMatV2 library (in Taichi) will take over and further back-propagate to a procedural parameters gradient. This requires passing gradients between multiple systems written in different languages.}


\paragraph{Texture descriptors}
Gatys et al.~\shortcite{Gatys2015} leverage a pre-trained VGG neural network~\cite{vgg19} to guide style transfer, using the Gram matrices of extracted deep features from the VGG layers as their statistical representation. We use 5 layers (the ones after each pooling operation in the VGG19 variant), and concatenate the flattened Gram matrices into a single vector descriptor of the crop texture. We use this loss to compare the texture content between image crops, and found it to be the most reliable of the alternatives. Note that the descriptor size does not depend on the input crop size, and can be used to compare differently-sized crops.

Heitz et al.~\shortcite{Heitz2021} introduced an alternative sliced Wasserstein loss, which in theory compares the distributions of VGG activations more accurately, though this is at the cost of introducing more noise into the gradients. Some previous works have generalized these descriptors to arbitrarily-shaped image regions given by masks \cite{photoscene, hu2022control}; however, the masks often need to be processed by case-dependent amounts of erosion, and we find that rectangular crop regions remain more reliable. 
\section{Method}
\label{sec:ours}

\begin{figure*}[t]
	\centering
	\includegraphics[width=1.0\textwidth]{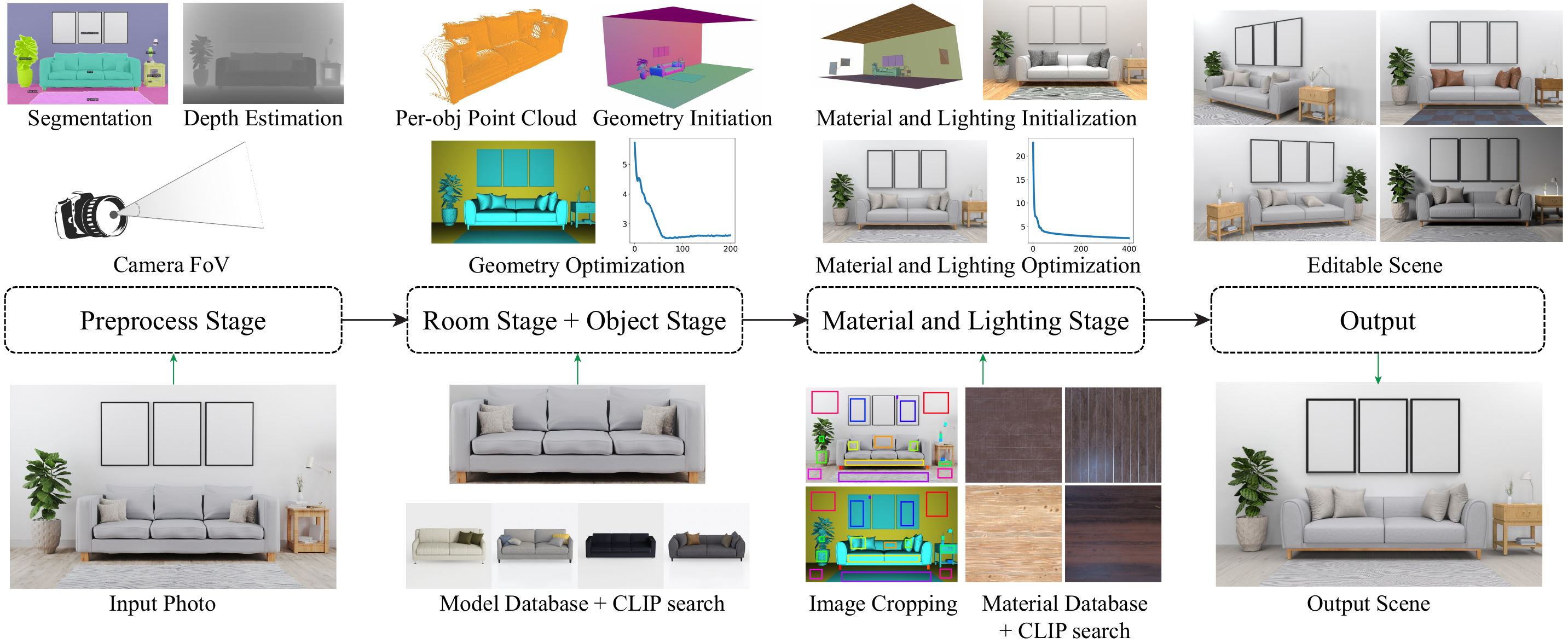} 
	\caption{\label{fig:fig_pipeline}
			A high-level overview of our pipeline. In the \emph{preprocess} stage (Sec. \ref{ssec:preprocess}), we segment the input image and estimate camera field-of-view and per-pixel depth. A user can optionally edit these estimates. In the \emph{room stage} (Sec. \ref{ssec:room}) and \emph{object stage} (Sec. \ref{ssec:object}), we initialize and optimize the geometry of the box representing the room, as well as the objects in it. The objects are selected from the 3D-Future database by search using similarity in CLIP space. Finally, the \emph{material and lighting} stage (Sec. \ref{ssec:finalstage}) selects procedural node graph materials from a database using CLIP similarity, initializes material texture transforms and lighting, and refines the estimates using optimization, making use of corresponding crop pairs between input and rendered images, which can be automatically detected or user-provided. The final output is a reconstructed scene, which supports camera, geometry, material and lighting edits.
	}
\end{figure*}

\paragraph{Overview}
Our \sysName system starts from a single input image of an indoor scene, with an object segmentation, depth and camera intrinsics (specifically field-of-view) provided by existing estimation methods.
Our goal is to obtain a visually matching scene reconstruction, which includes picking the right objects, optimizing their poses, picking the right materials, and optimizing the lighting and materials in the scene. For most objects (at least the ones where texturing makes sense) our goal is to obtain tileable material textures, including diffuse color, roughness and normal maps.

Our method includes four high-level stages. First, in the \emph{preprocess stage} we predict the camera field-of-view and estimate the depth map and segmentation per object; this stage is entirely based on previous methods. These inputs can be interpreted as an approximate segmented point cloud, which is sufficient for initialization decisions. Second, in the \emph{room stage}, we initialize a coarsely aligned room box based on the subset of the point cloud belonging to the room's walls, floor and ceiling, and use differentiable rendering to refine the room box to satisfy an image-space mask loss in combination with a depth loss. Next, in the \emph{object stage}, we search a database of objects to match the segmented objects in the input image, coarsely align them to their point clouds and use differentiable rendering to refine their poses to minimize their image-space mask loss and depth loss. Finally, in the \emph{material and lighting stage}, we choose either a homogeneous material or a procedural node graph from a database for each material part, initialize lighting in the scene, and use differentiable rendering to jointly optimize the materials and lighting; for procedural node graph materials, this optimization backpropagates to their node parameters.

\subsection{Preprocess stage} \label{ssec:preprocess}
\paragraph{Camera intrinsics}
We assume a simple pinhole camera model. Given an input image, we use \rev{PerspectiveFields~\cite{perspectivefields} to estimate the camera vertical field-of-view and assume square pixels. In our experiments, the model is robust enough and its predictions do not need manual adjustments.}
\paragraph{Depth estimation}
We use a recent monocular depth estimation proprietary model based on DPT~\cite{ranftl2021vision} and MiDaS~\cite{ranftl2020towards}, to generate an approximate depth map. As the depth units returned are unknown, we simply used the normalised depth and do not assume anything about absolute scale. While not perfect, this depth information works reasonably well for room and object placement, which we will further refine using differentiable rendering, considering image-space losses. We can use depth information from any other source, e.g. a depth sensor. 

\paragraph{Segmentation}
We use Mask2Former~\cite{Mask2Former} to perform a panoptic segmentation, which yields a mask image for each object, including the room ceiling, floor and walls; these are labeled by the model, which we can use to establish correspondence with our rendered box and objects. 

\begin{figure}[t]
	\centering
        \small
        \begin{tabular}{cc}
        Initialization & Optimized Result \\
        \includegraphics[height=0.95in]{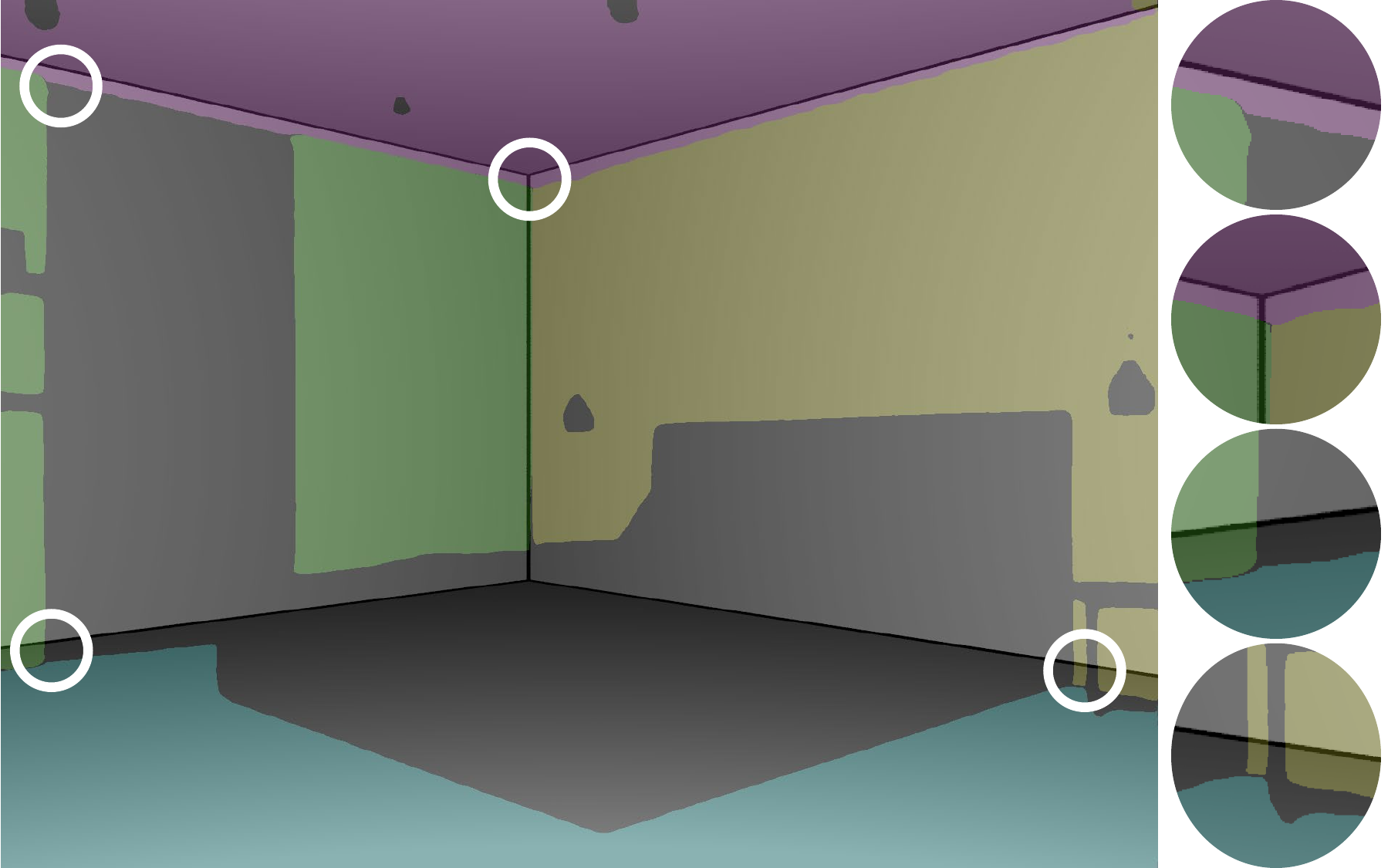} &
        \includegraphics[height=0.95in]{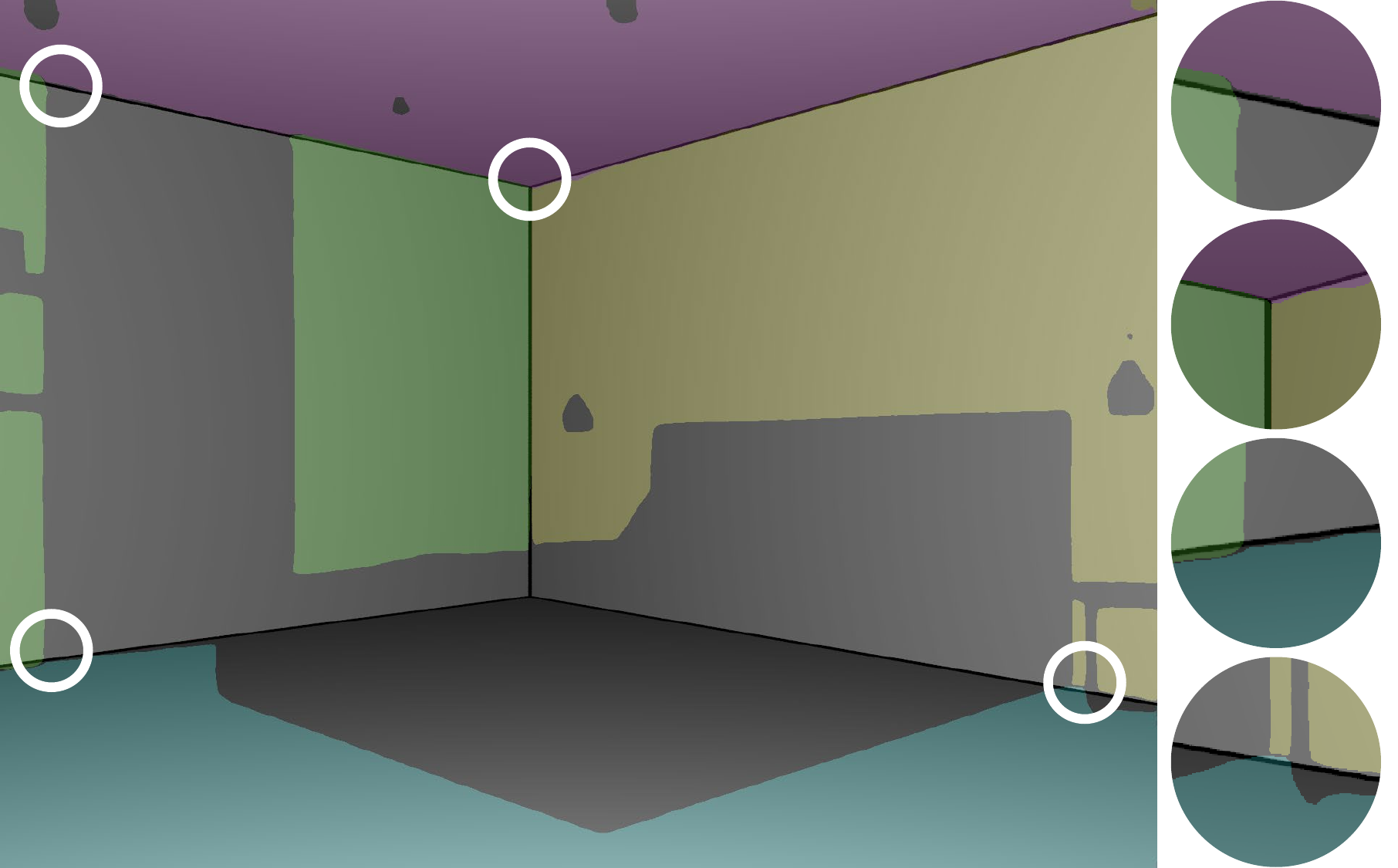}
        \end{tabular}
	\caption{
		 Room stage optimization: We show the estimated segmentation masks of the walls/floor/ceiling versus the rendered room box before and after the room optimization. The optimized room box is better aligned to the mask (see circular insets). Also note that the estimated masks are much smaller than the rendered box sides, as the room was not empty; our differentiable IoU metric is appropriate for this scenario.}
 \label{fig:RoomStage}
\end{figure}

\subsection{Room stage} \label{ssec:room}

For simplicity we assume the walls, ceiling and floor of the room can be approximated as a box; more complex floor plans could be supported in the future using an extension of our approach. We aim to initialize a coarsely aligned room box to the approximate point cloud estimated above, followed by a refinement stage, where we further optimize the box size and placement to match the image-space segmentation.

\paragraph{Room initialization}
In the coarse stage, we take the depth and segmentation masks corresponding to room walls, ceiling and floor (with some erosion applied), to generate a segmented room point cloud, without objects. We estimate the floor plane using  RANSAC on the floor segmented point cloud. We then estimate the room height through maximum height point value. If the room does not have a ceiling visible in the image, we simply set the room height sufficiently above the top visible point (1.2$\times$maximum height point value). We rotate the room box horizontally around the predicted floor normal and find the rotation+scaling aligning best with the RANSAC estimated wall from the point cloud. This results in a coarsely aligned room box that approximately matches the predicted room point cloud, and is sufficient as initialization. 

\paragraph{Room optimization}
We next use differentiable rendering to further optimize the room box, improving visual alignment to the input image. We optimize the pose (object-to-world matrix) of the room, including rotation, translation and scaling.
We base our differentiable renderer on \psdrcuda and optimise the walls, floor and ceiling positions based on depth maps and predicted maps differences. As \psdrcuda handles the edges discontinuities defined in Section~\ref{ssec:prelim_psdr}, losses based on depth/mask can be differentiated and optimized smoothly. Note that the rendered masks have anti-aliased edges, which is crucial for gradient descent to work, as only the edge pixels will have non-zero gradients. We use the predicted depth and masks to compute our loss function, minimizing the mean L1 difference of depth predictions, and maximize the intersection over union (IoU) for room walls, ceiling and floor. The IoU metric is extended to correctly handle fractional anti-aliased pixels, giving correct gradients from these terms. 
\begin{equation}
    \loss_{box} = -\sum_{i} IoU(M^r_i(\theta_{box}), M^t_i) + \loss_1(D_r(\theta_{box}), D_i),
\end{equation}
where $\theta_{box}$ are the room "box" parameters, $i$ varies over the set \{wall, floor, ceiling\}, $M^r_i$ and $M^t_i$ are the rendered and target masks respectively, and $D^r$ and $D^t$ are the rendered and target depths respectively. We normalize depths to a unit range.

\subsection{Object stage} \label{ssec:object}

\paragraph{Model database}
We use the 3D-FUTURE dataset~\cite{3d-future} as the 3D model database. It contains a total of 10,000+ artist-made models (mostly furniture and similar assets typical in indoor scenes) with split material groups for each model, so that objects with multiple materials are possible (for example, a sofa with pillows of a different material). We discard the materials provided by the dataset, since we want to retrieve and optimize materials depending on the input photograph.

\begin{figure}[t]
    \centering
    \small
    \includegraphics[height=1.3in]{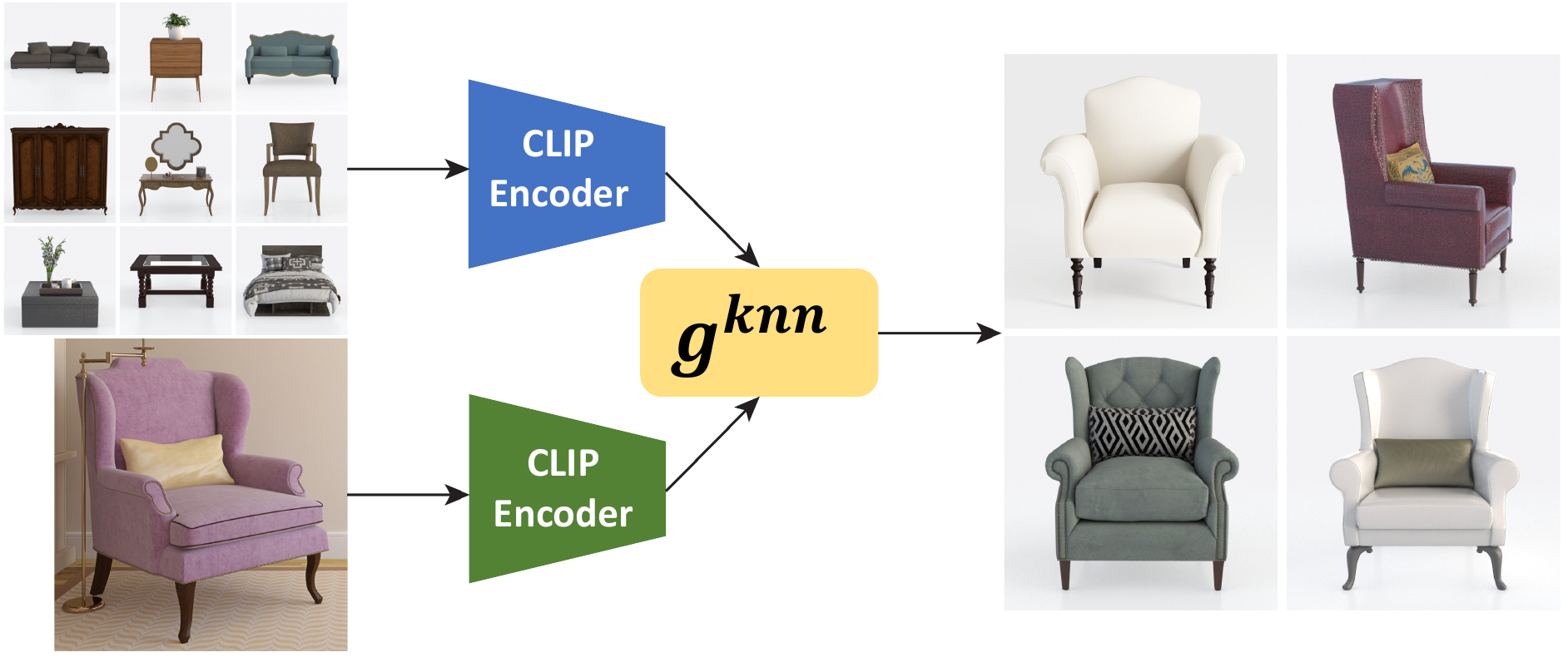}
    \caption{\label{fig:CLIP_Model}
        We use CLIP as a zero-shot ranked classifier of 3D geometries from the 3D-Future database. We encode the renderings provided with the database as well as a crop around each object from our input image, and find nearest neighbors using cosine similarity.
    }
\end{figure}

\paragraph{Model search}
We use CLIP~\cite{CLIP} to search 3D models from the database that match the input scene objects.
We use the renderings available in the database and encode them all using CLIP into normalized 768-dimensional vectors.
To search an appropriate model for each scene object in our segmentation, we crop a target image around each object's mask and encode the cropped image using CLIP. We then select the Nearest Neighbors --using a cosine distance- in CLIP space from the database, providing the closest assets in the database matching the input crop. 

\paragraph{Coarse geometry prediction}
Similarly to the room stage, we predict each object's pose by fitting to the approximate point cloud of the object and refining it using 
gradient-based optimization to match the image-space segmentation.
For initial alignment, we first find the center as the median value of the point cloud over the three axes. We then compute the scaling factor by matching the mesh bounding box center to the point cloud median, and the mesh radius to half the point cloud radius, where radius is defined as median distance to center. While this process initializes the scaling to a smaller value than reality, it provides a good starting point for the later stage of geometry optimization; making the objects too small at initialization and letting them grow appears to make the optimization better behaved than trying to initialize the scale accurately.

We make the vertical axis for each object orthogonal to the room floor; this heuristic is appropriate for most objects common in indoor scene settings. We try a number of rotations around the vertical axis, which is kept orthogonal to the room floor. 
We pick as initial rotation the object rotation which has the minimum mean L1 difference between the rendered mask and object segmented mask.
If the bottom face of an object's bounding box and the floor are close (distance < 0.1) and the mask edge is close enough to the floor mask (distance < 20px), we mark the object as ``on the floor'' and use a floor-distance loss when optimizing that object's position. \revky{If an object is floating, we search if the object mask edge is close enough (distance < 20px) of another object edge on the floor. If so, we enforce the floating object to stay on top of the parent object.}

\paragraph{Refining geometry}
Next, we use differentiable rendering to perform a joint pose optimization for all objects. We optimize the scaling, horizontal translation and rotation around the vertical axis --keeping alignment with the room floor/ceiling--for each object. This ensures that objects snapped to the floor/ceiling preserve this constraint. Similarly, for objects on the wall, the optimization transforms and scales them only along the wall. Our differentiable renderer generates the masks with anti-aliasing, which is critical for non-zero gradient in masks comparing losses. \revky{For the loss function, we use the mean L1 difference between the rendered and estimated segmentation mask multiplied by the depth of each object. We also use the L1 difference of heavily blurred versions of the masks, using a Gaussian blur.}

\begin{multline}
    \loss_{obj} = \sum_{i} \loss_1(M^r_i(\theta_{obj})*D_r(\theta_{obj}), M^t_i*D_i) \\
    + \sum_{i} \loss_1(G(M^r_i(\theta_{obj}), G(M^t_i))),
\end{multline}
\revky{where $\theta_{obj}$ are the object pose parameters, $i$ varies over the set of objects, $M^r_i$ and $M^t_i$ are the rendered and target masks respectively, and $D^r$ and $D^t$ are the rendered and target depths respectively, normalized to unit range. $G$ is a 2D Gaussian convolution pyramid of a mask.}

\begin{figure}[t]
    \centering
    \begin{tabular}{c}
        \includegraphics[height=1.4in]{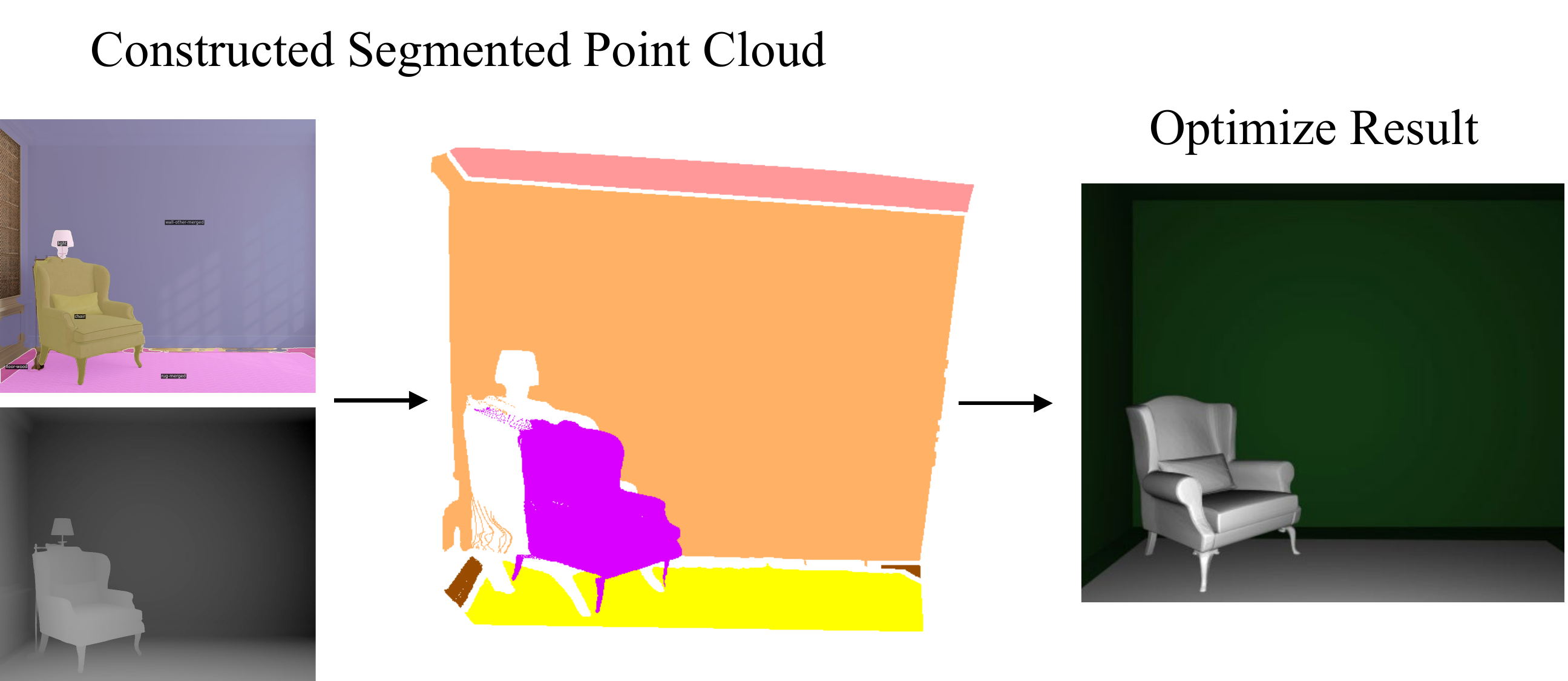}
    \end{tabular}
    \caption{\label{fig:Geo_Stage}
        Object Stage: We take the segmentation and depth map to construct segmented point clouds for coarse geometry prediction. We initialize object poses to the point cloud and use differentiable rendering in image space (with mask and depth losses) to further refine the poses.
    }
\end{figure}

\subsection{Material and lighting stage} \label{ssec:finalstage}
In this stage, we predict and optimize the lighting and materials for the scene. For this, we establish a list of \emph{crop pairs} between the input image and the rendering. Our optimization will match appearance between the crops of each pair, in addition to optimizing a low-resolution full-image loss. These crop pairs are normally automatically detected, but can optionally be provided by the user for best appearance match.

\paragraph{Crop pairs}

To achieve the best robustness and closest appearance match for different material parts, we sample a number of crops inside the intersection region of a mesh material group rendered mask and the object segmentation mask. 
\rev{%
    For each object $i$, let $M_i \defeq M_i^r(\theta_{obj}) \cap M_i^t$ be the intersection between the rendered $M_i^r$ and the target $M_i^t$ masks for this object.
    Then, we compute a rectangular mask $W_i \subset M_i$ with maximal size to use as the crop window for this object in both the render and the target images.
    In this way, $(I_i, R_i)$ forms a crop pair where $R_i = R[W_i]$ and $I_i = I[W_i]$ denote, respectively, crops of the rendered image $R$ and the input image $I$ using the  window $W_i$.
}


To compare the textures between each pair of crops during optimization, we use a Gram matrix loss as we found it very reliable for optimization, outperforming alternative losses like masked VGG.

\paragraph{Light initialization}
In this stage, we generate the initial lighting for the scene.
We split the room ceiling into a grid of area lights and \revky{scale them down by a ratio of 0.8}. We remove the lights that intersect the camera frustum, or are completely behind the camera. We also add one large  area light behind the camera and one per invisible wall, simulating an invisible window. \revky{We also turn visible windows and lamps into emissive objects. For windows, we add a simple frame, though a database of window geometries could be used if available. For lamps, we make the largest visible rendered part into emissive.}

We try a number of uniform light radiance values as initialization and pick the one with the minimum down-sampled L1 loss on luminance images as our initial light radiance. As light transport is linear in light intensity, a single rendering with unit radiance is needed for this step.

\paragraph{Material initialization and search}
\revky{For material parts that do not have any available crop pair because of small mask area, or the cosine similarity for the best matched CLIP search result is worse than 0.25, we use homogeneous parameters (constant albedo, roughness, specular). We initialize homogeneous material albedo to the median color of the mesh-mask intersection area and a roughness of 0.5. If there is no intersection, we use the median color of the area of the entire object segment.

For each material part that needs a procedural material (i.e. that has at least one crop pair $(I_i, R_i)$), we use the crop $I_i$ to run a CLIP search on rendered thumbnails of 118 procedural materials from the DiffMatV2 library provided by the MATch method \cite{li2023diffmatv2}.  Our material search is based on cosine similarity of normalized CLIP encodings, same as the object search described above. If multiple crops are on a single material part, all pairs are used for CLIP search; for each crop we evaluate the top-10 similar materials. We use a voting scheme and select materials appearing for multiple crops. Out of these selected materials, we compare cosine similarities and select the material with highest mean cosine similarity.}
To determine the right texture transform, we sample the chosen material with different scales (0.5-8.0) and rotations (-45, 0, 45, 90) and find the one with minimum Gram matrix loss on that crop pair. If the Gram matrix loss of a homogeneous material is better than any of the texture transforms, we refrain from applying textures to this material and treat it as homogeneous. Note that we render and compare these crops under the initial lighting in the scene a the final lighting has not yet been optimized at this stage.

\paragraph{Full scene material and lighting optimization}

\revky{In the final stage of our pipeline, we jointly optimize all materials and lights in the scene, under full global illumination. The loss function we use combines several terms: the L1 difference between the input image and rendering (downsized to 1/8), the Gram matrix difference for each crop pair, as well as the mean RGB color per object mask:}
The complete loss for joint material and lighting optimization is as follows:
\begin{multline}
    \loss_{final} = \loss_1(R_{1/8}(\theta), I_{1/8}) \ + \\ \sum_{i}  \loss_1(\mu(M_i^r(\theta_{obj})), C_i) +  \sum_{i} \loss_1(T_G(R_i(\theta)), T_G(I_i)) ,
\end{multline}
\revky{where $\theta$ is the vector of all material and lighting parameters, $R_{1/8}$ and $I_{1/8}$ are the rendered and input image, downsized to 1/8 of original size of height and width, $i$ iterates over all crop pairs, $(I_i, R_i)$ are the crop pairs, $\mu$ denotes the RGB mean over a mesh mask, $C_i$ denotes the median color computed during initialization and $T_G$ is the Gram matrix texture descriptor of a crop. 
Note that the optimization over procedural material parameters is bound to a plausible texture manifold allowed by the node graph, acting as a prior and ensuring the optimization generates only high-quality textures without baking in lighting cues, even in invisible parts of the scene.}

\section{Results}
\label{sec:results}
%

\begin{figure}[t]
    \centering
    \small
    \setlength{\resLen}{.925in}
    \addtolength{\tabcolsep}{-3.5pt}
    \begin{tabular}{cc}
        \begin{overpic}[trim=0 0 0 170,clip,height=\resLen]{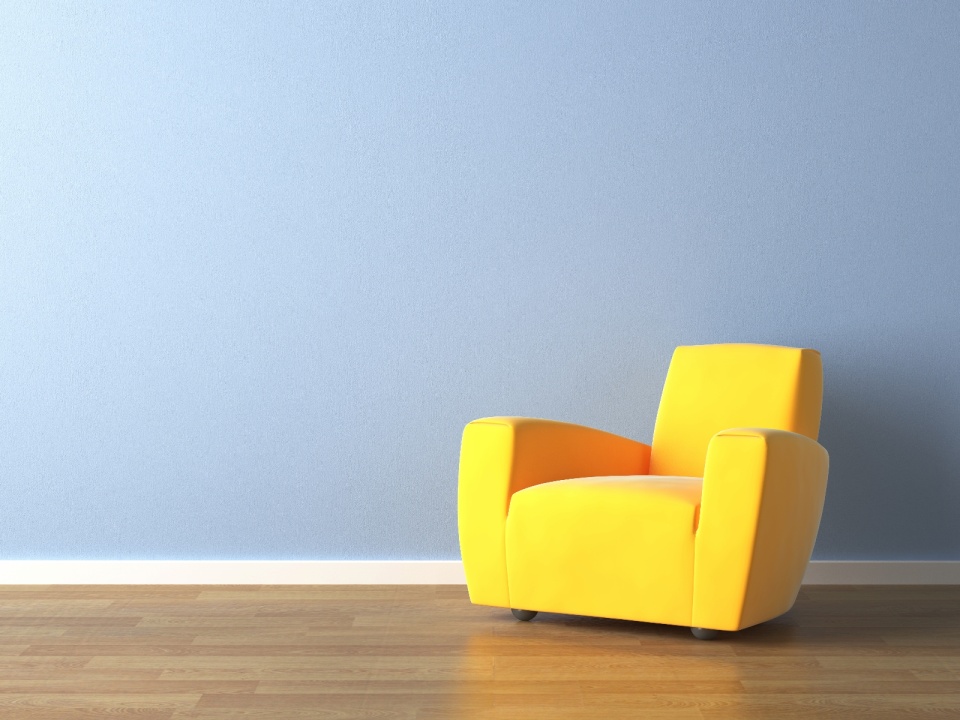}
            \put(2, 50){\color{white} \contour{black}{(a) \textbf{Input photo}}}
        \end{overpic}
        &
        \begin{overpic}[trim=0 0 0 170,clip,height=\resLen]{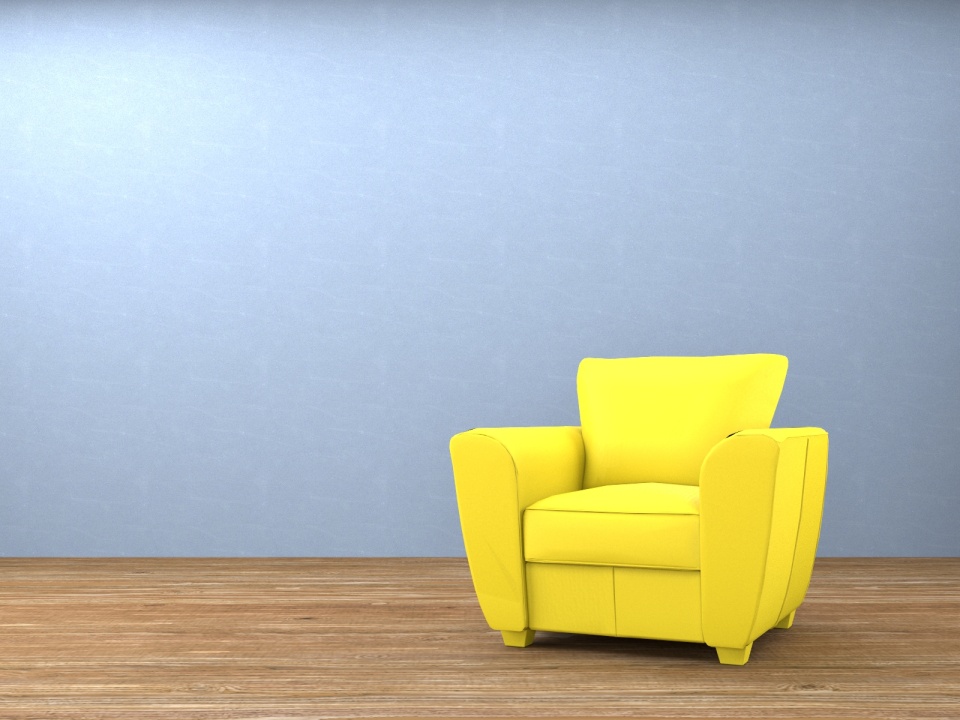}
            \put(2, 50){\color{white} \contour{black}{(b) \textbf{Ours}}}
        \end{overpic}
        \\
        \begin{overpic}[trim=0 0 0 170,clip,height=\resLen]{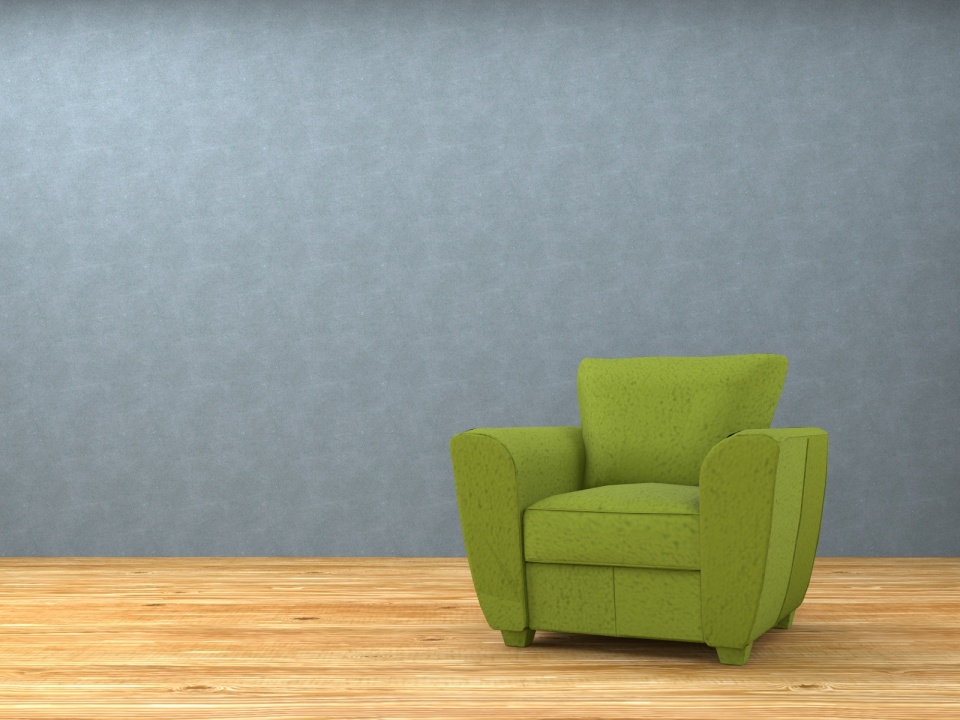}
            \put(2, 50){\color{white} \contour{black}{(c) \textbf{Perturbed params.}}}
            \end{overpic}
            &
        \begin{overpic}[trim=0 0 0 170,clip,height=\resLen]{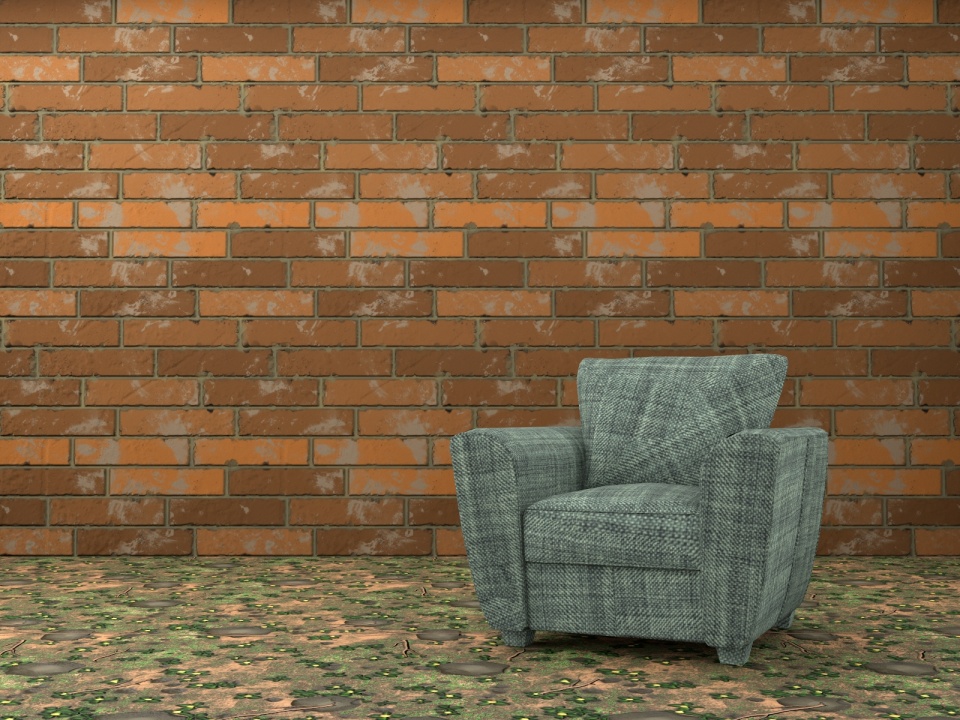} 
            \put(2, 50){\color{white} \contour{black}{(d) \textbf{Different graph}}}
        \end{overpic}
    \end{tabular}
    \caption{\label{fig:fig_edit_mat}
        Examples of material variations. We can modify the parameters of the procedural node graphs (c) or switch to a completely different procedural graph material for the floor (d).
    }
\end{figure}

\paragraph*{Scene edits}
%
%
In Fig. \ref{fig:fig_edit_mat}, we show examples of material editing.
We can 
modify the parameters of the procedural node graphs (c) or switch to a completely different procedural graph material for the floor (d).
\rev{Please refer to the Supplemental Material for more editing examples.}

\begin{figure}[t]
    \centering
    \small
    \setlength{\resLen}{1.02in}
    \addtolength{\tabcolsep}{-4pt}
    \begin{tabular}{cc}
        \begin{overpic}[height=\resLen]{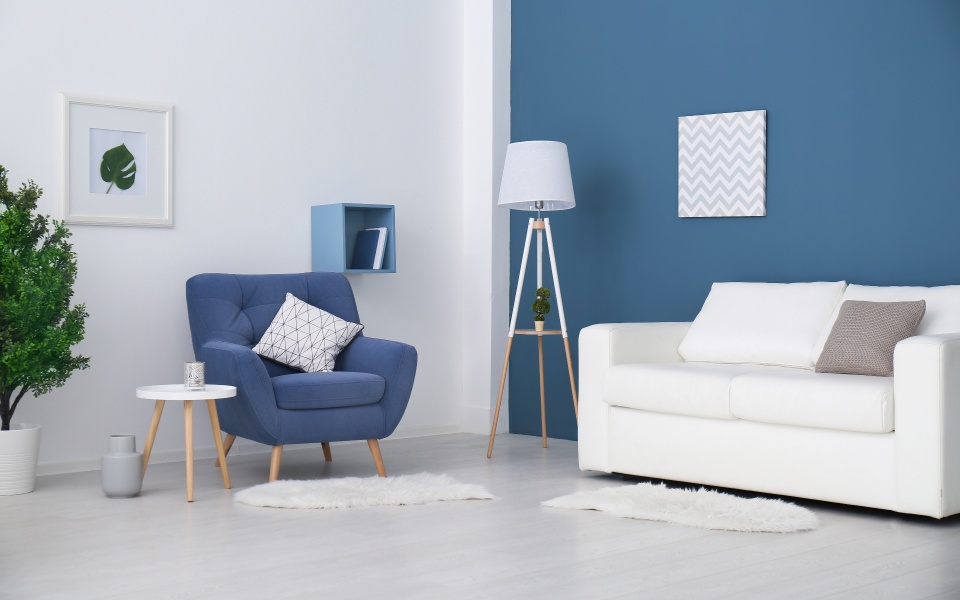}
            \put(2, 4){\color{white} \contour{black}{(a) \textbf{Target}}}
        \end{overpic}
        &
        \includegraphics[height=\resLen]{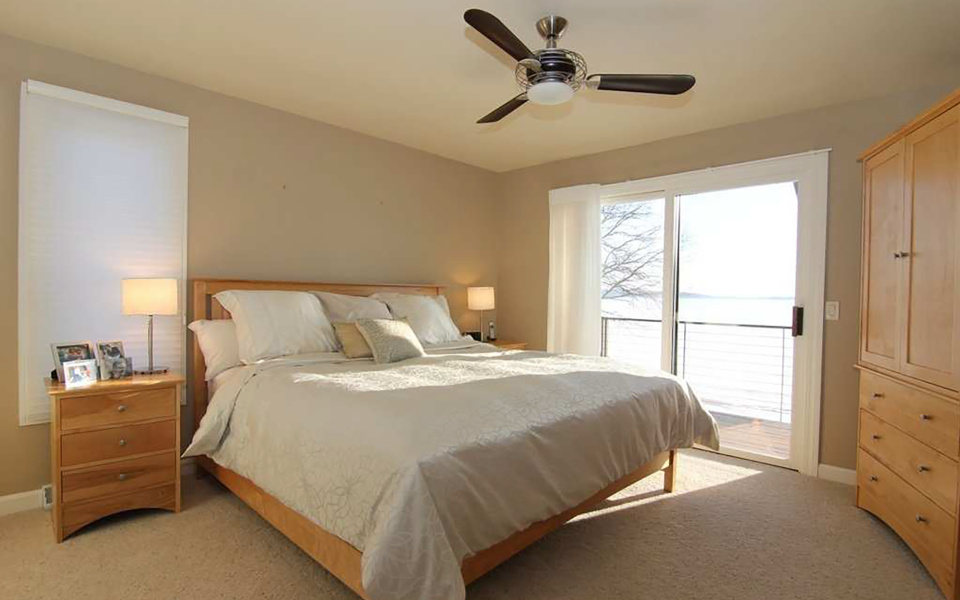}
        \\
        \begin{overpic}[height=\resLen]{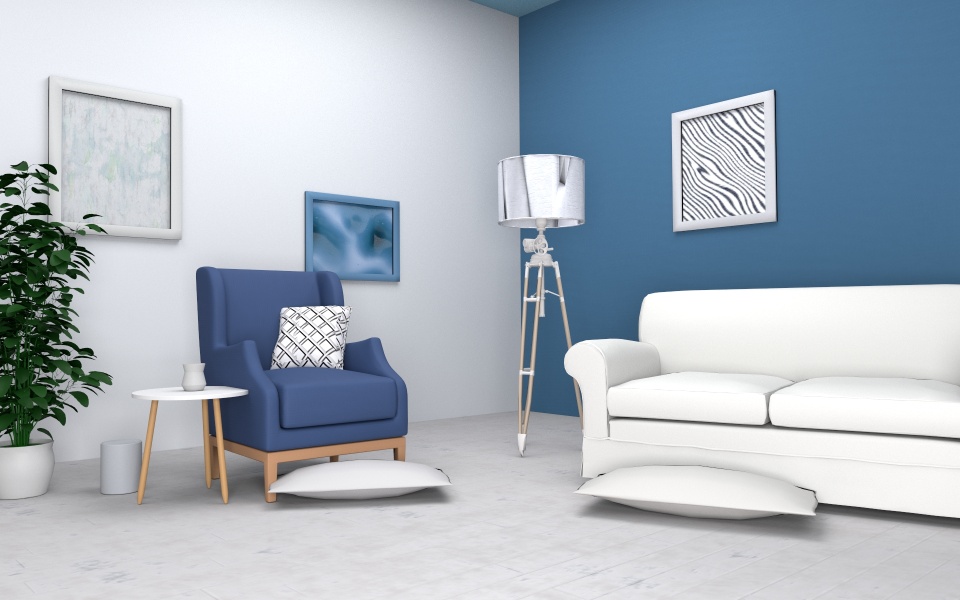}
            \put(2, 4){\color{white} \contour{black}{(b) \textbf{Ours} ($100\%$)}}
        \end{overpic}
        &
        \includegraphics[height=\resLen]{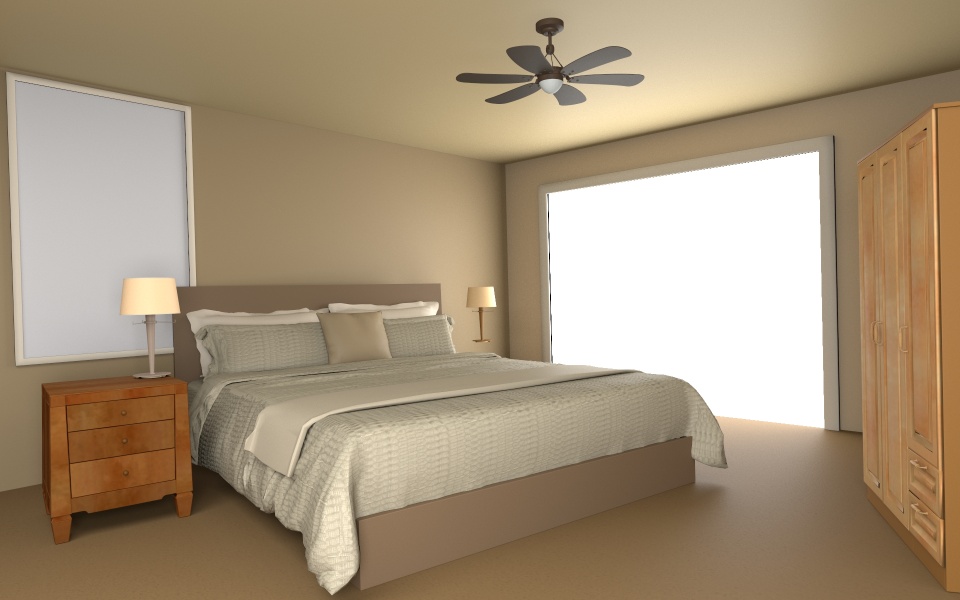}
        \\
        \begin{overpic}[height=\resLen]{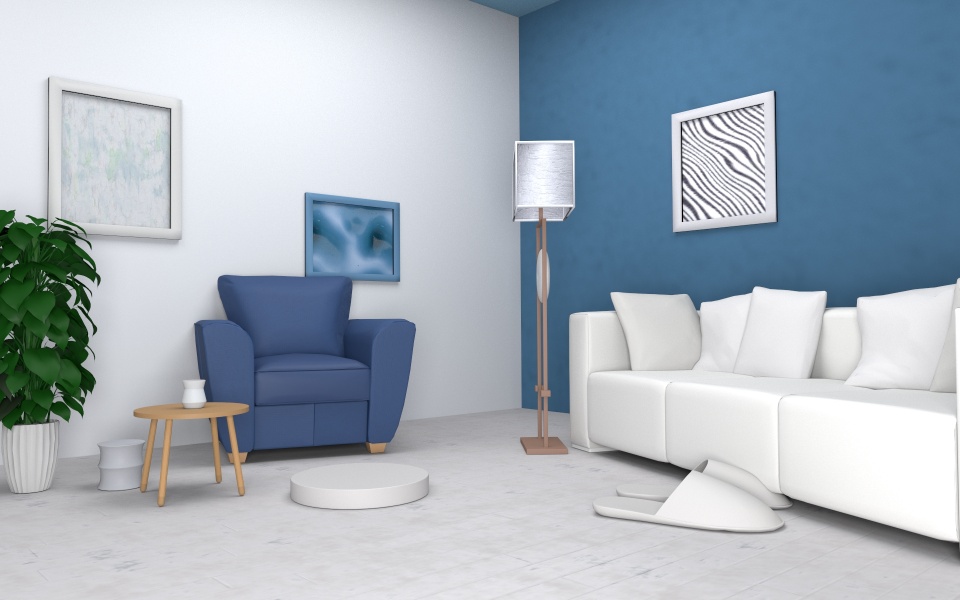}
            \put(2, 4){\color{white} \contour{black}{(c) \textbf{Ours} ($5\%$)}}
        \end{overpic}
        &
        \includegraphics[height=\resLen]{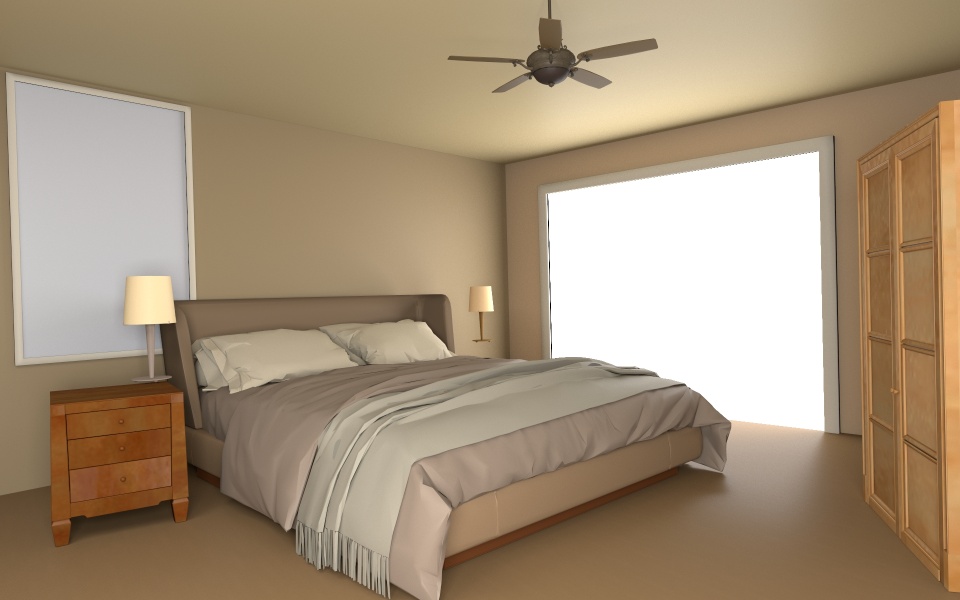}
    \end{tabular}
    \caption{\label{fig:db_size}\rev{
        Our pipeline is robust to the choice of shape databases.
        Rows (b) and (c) show our results for two example target images generated using, respectively, a full database (with 16563 models) and one containing $5\%$ of randomly sampled data.
    }}
\end{figure}

\rev{
    \paragraph*{Ablation on database size}
    As described in \S\ref{sec:ours}, our pipeline retrieves object geometries from an input database.
    Fortunately, as demonstrated in Fig.~\ref{fig:db_size}, our pipeline is robust to the choice of this database.
    Even with only $5\%$ of the data used (containing 800 3D models), our pipeline remains well-behaved.
}

\begin{figure}[t]
    \centering
    \small
    \setlength{\resLen}{1.1in}
    \addtolength{\tabcolsep}{-4pt}
    \begin{tabular}{cc}
        \begin{overpic}[height=\resLen]{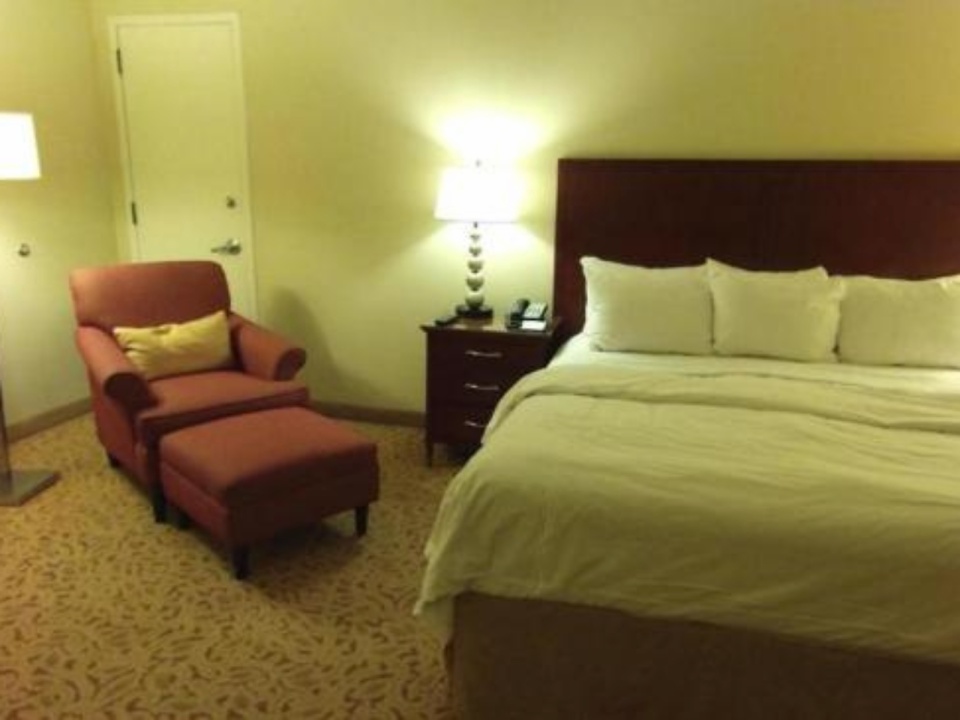}
            \put(2, 4){\color{white} \contour{black}{(a) \textbf{Target}}}
        \end{overpic}
        &
        \includegraphics[height=\resLen]{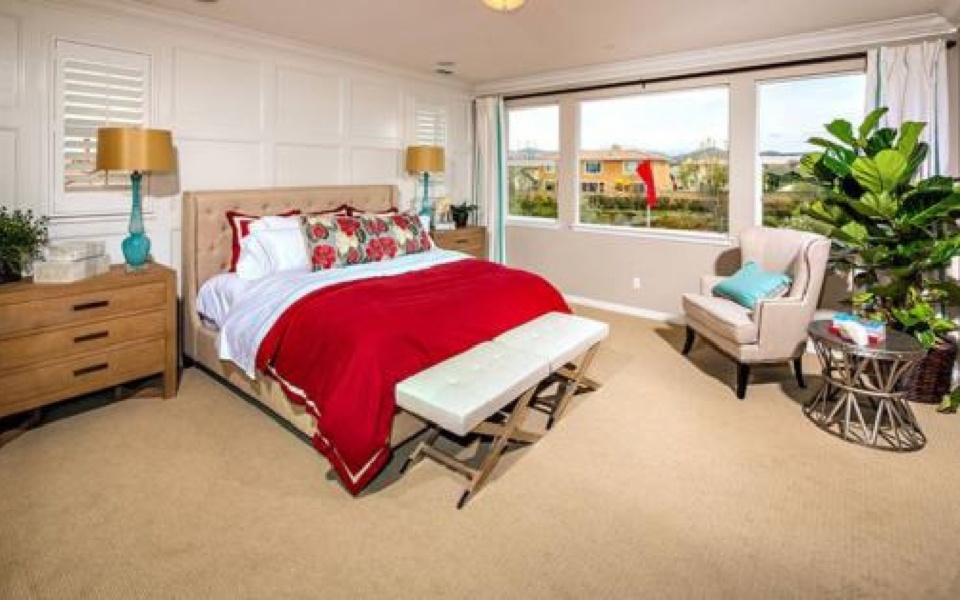}
        \\
        \begin{overpic}[height=\resLen]{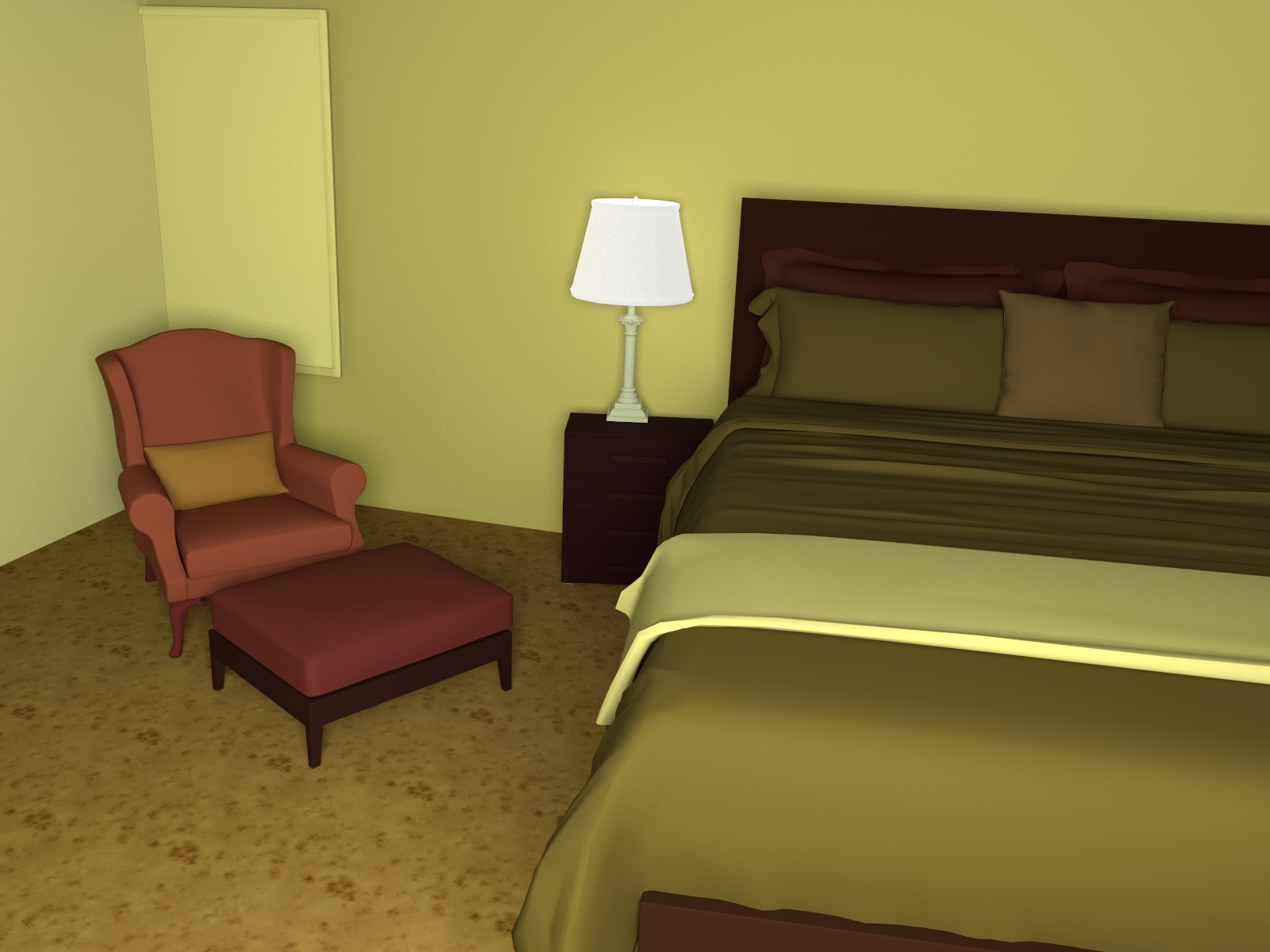}
            \put(2, 4){\color{white} \contour{black}{(b) \textbf{Ours}}}
        \end{overpic}
        &
        \includegraphics[height=\resLen]{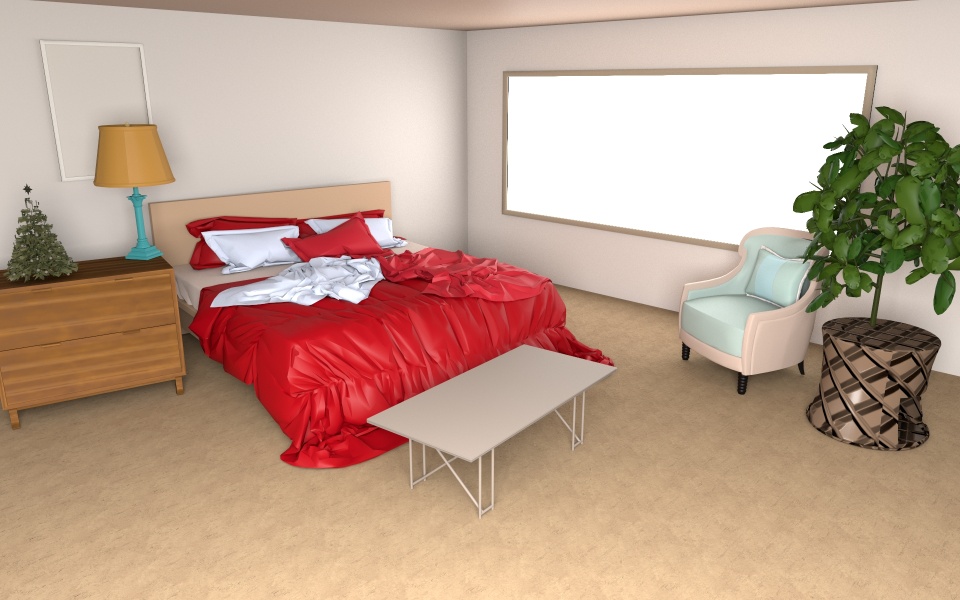}
        \\[-2pt]
        \emph{LPIPS \& RMSE:} \textbf{0.481} / \textbf{0.038} & \textbf{0.598} / \textbf{0.039}
        \\[2pt]
        \begin{overpic}[width=1.3334\resLen,height=\resLen]{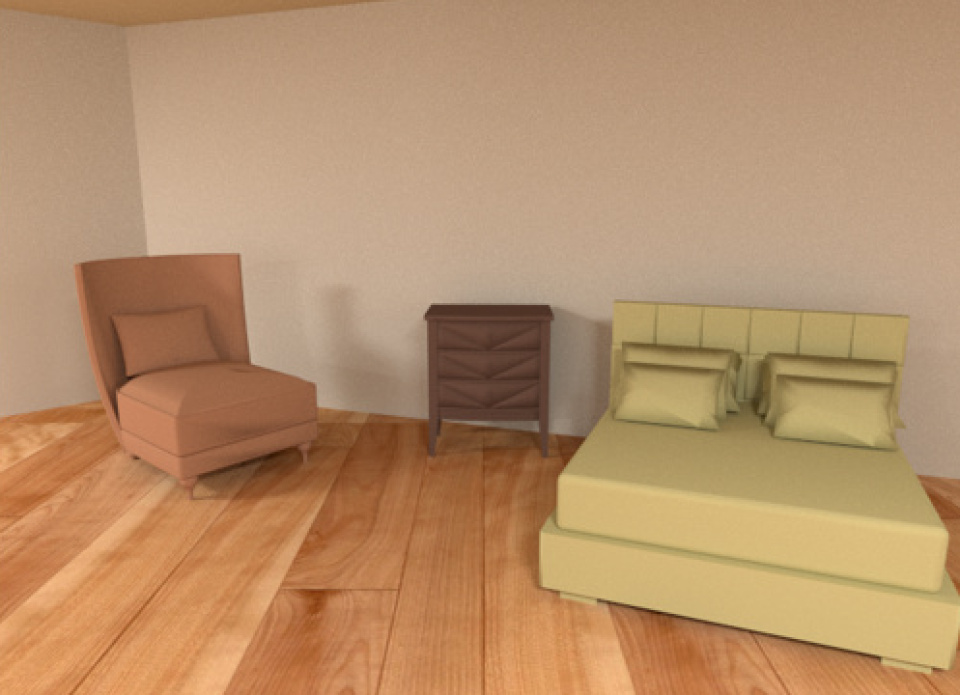}
            \put(2, 4){\color{white} \contour{black}{(c) \textbf{IM2CAD}}}
        \end{overpic}
        &
        \includegraphics[height=\resLen]{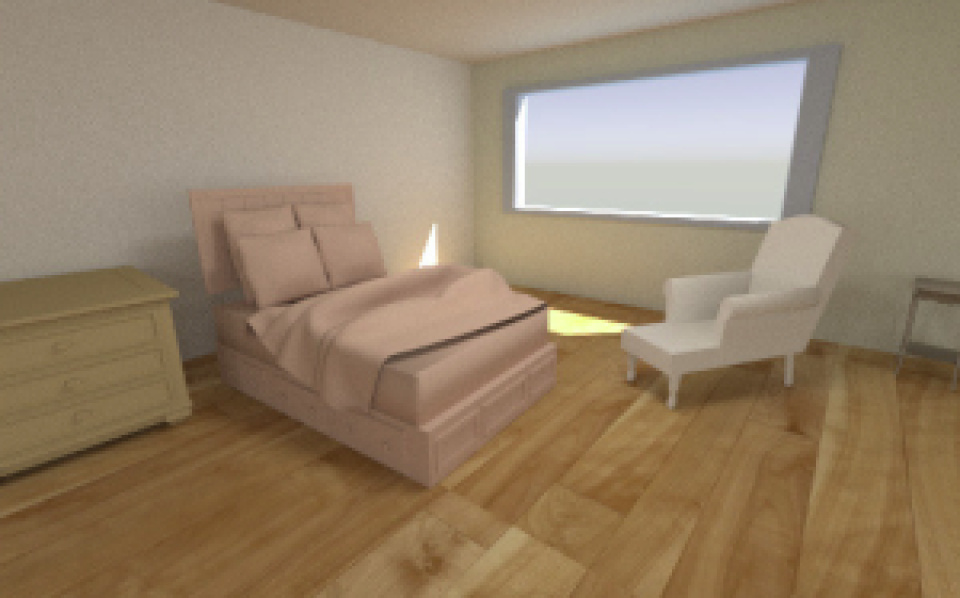}
        \\[-2pt]
        \emph{LPIPS \& RMSE:} 0.636 / 0.040 & 0.680 / 0.041
    \end{tabular}
    \caption{\label{fig:compare_im2cad}\rev{
        \textbf{Comparison with \textsf{IM2CAD}:}
        We compare results generated using our (automated) pipeline and \textsf{IM2CAD} using examples from their paper~\cite{im2cad}. 
    }}
\end{figure}

\begin{figure*}[t]
    \centering
    \small
    \setlength{\resLen}{1.4in}
    \addtolength{\tabcolsep}{-4.5pt}
    \begin{tabular}{ccccc}
        \textbf{Example 1} & \textbf{Example 2} & \textbf{Example 3} & \textbf{Example 4} & \textbf{Example 5}
        \\
        \begin{overpic}[width=\resLen]{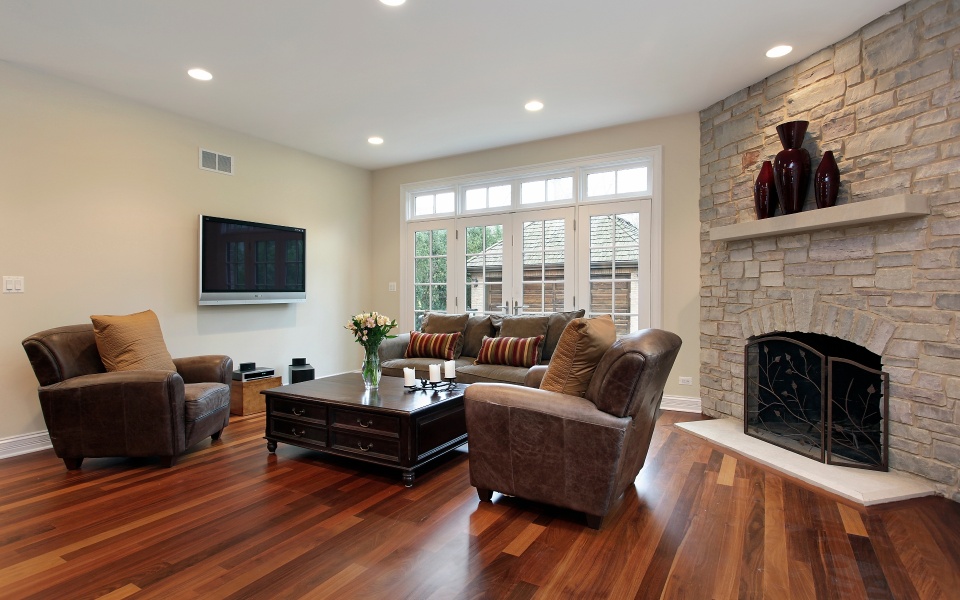}
            \put(2, 4){\color{white} \contour{black}{(a) \textbf{Target}}}
        \end{overpic}
        &
        \includegraphics[width=\resLen]{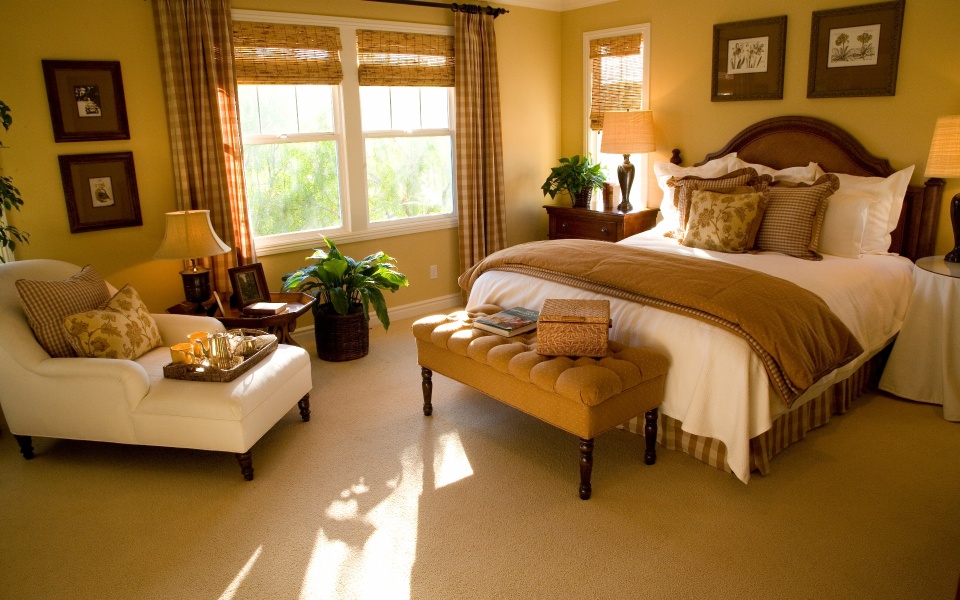} &
        \includegraphics[width=\resLen]{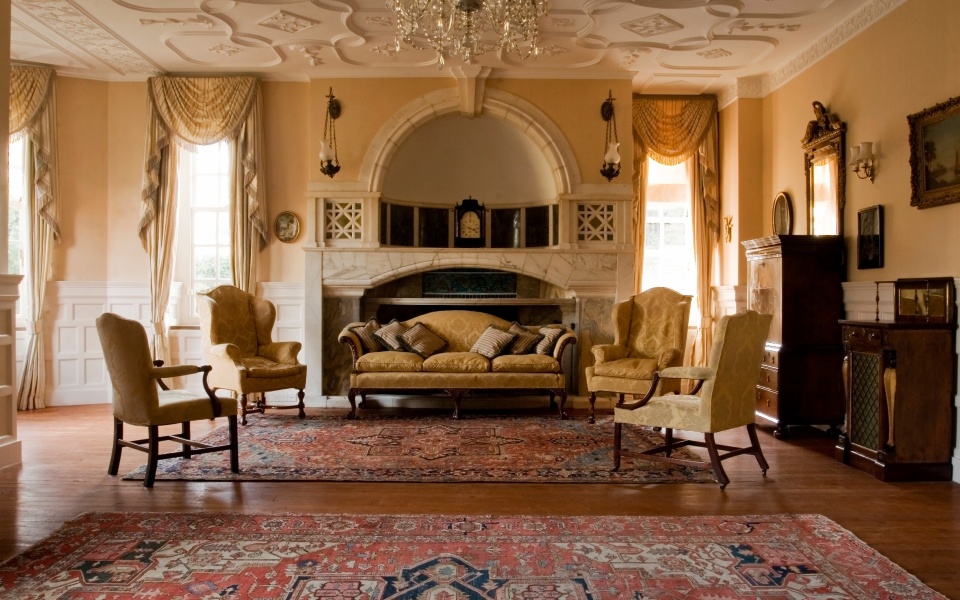} &
        \includegraphics[width=\resLen]{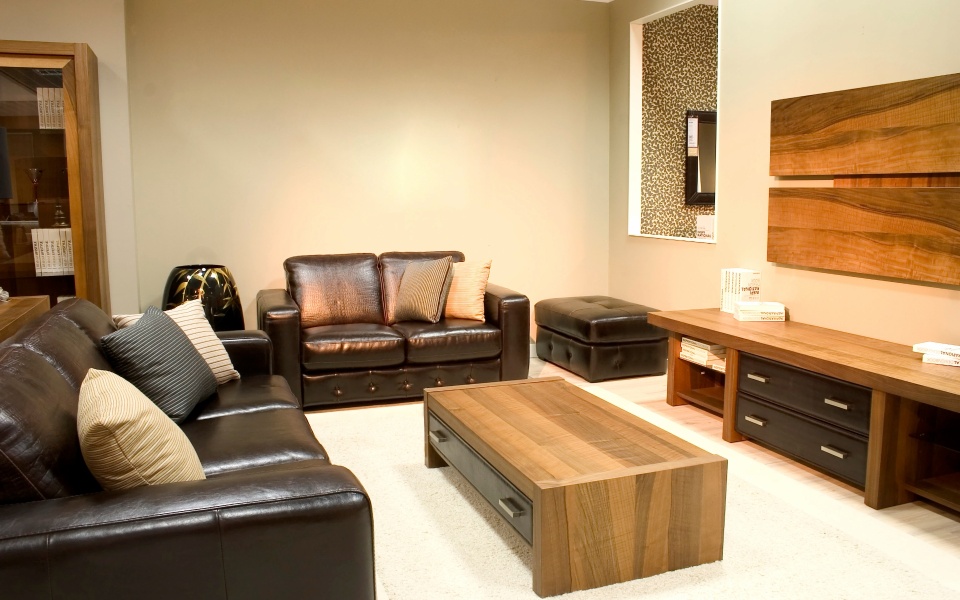} &
        \includegraphics[width=\resLen]{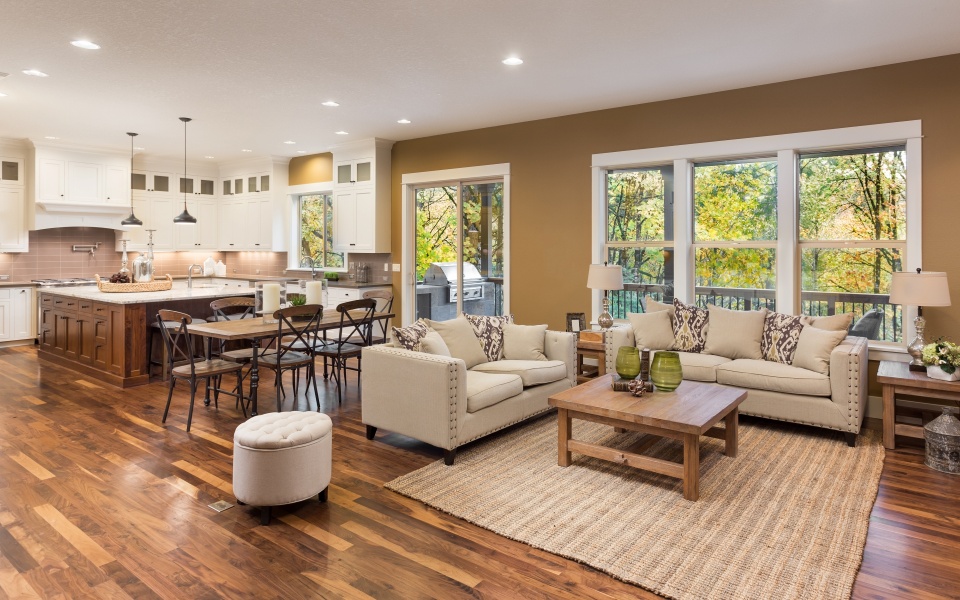}
        \\
        \begin{overpic}[width=\resLen]{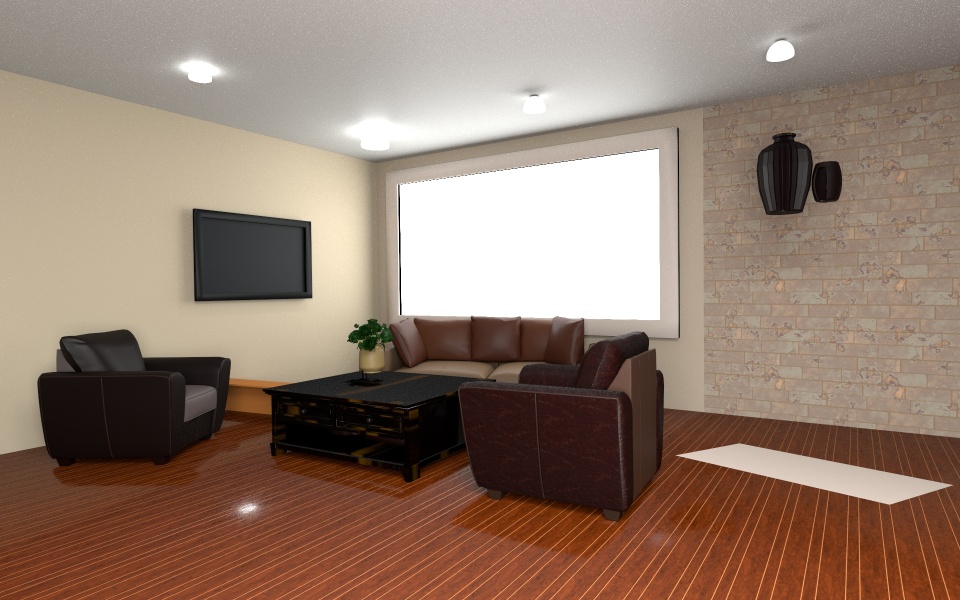}
            \put(2, 4){\color{white} \contour{black}{(b) \textbf{Ours}}}
        \end{overpic}
        &
        \includegraphics[width=\resLen]{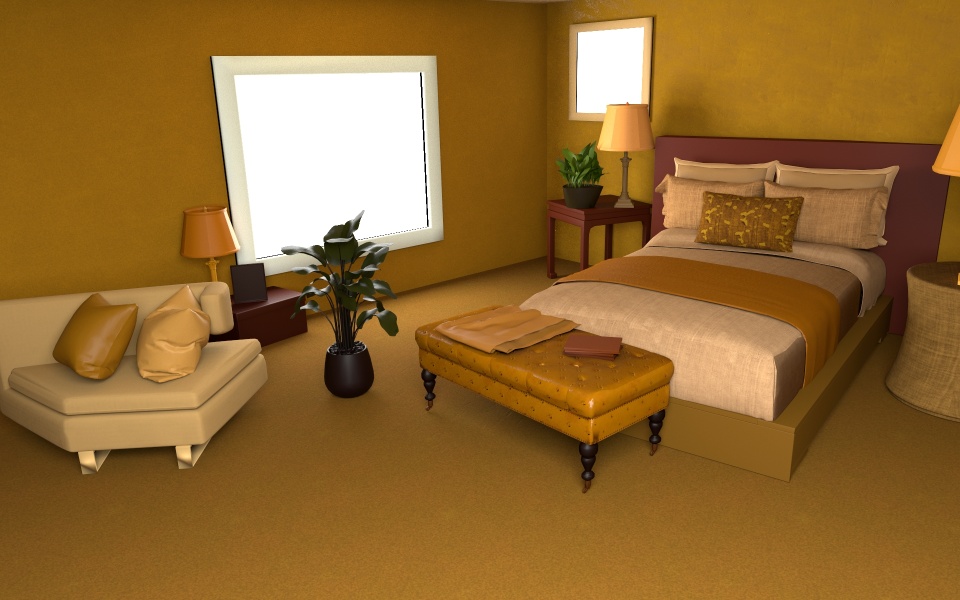} &
        \includegraphics[width=\resLen]{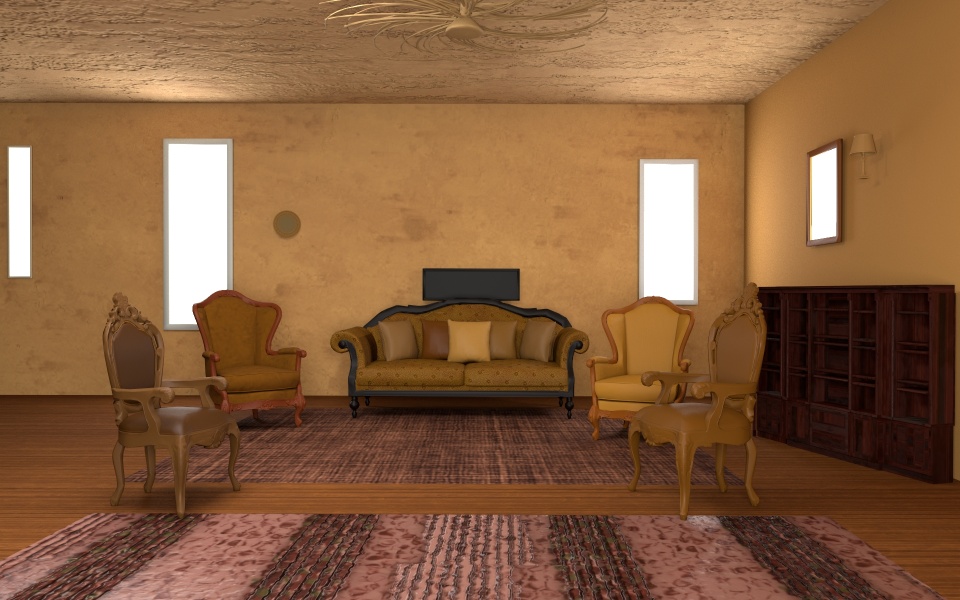} &
        \includegraphics[width=\resLen]{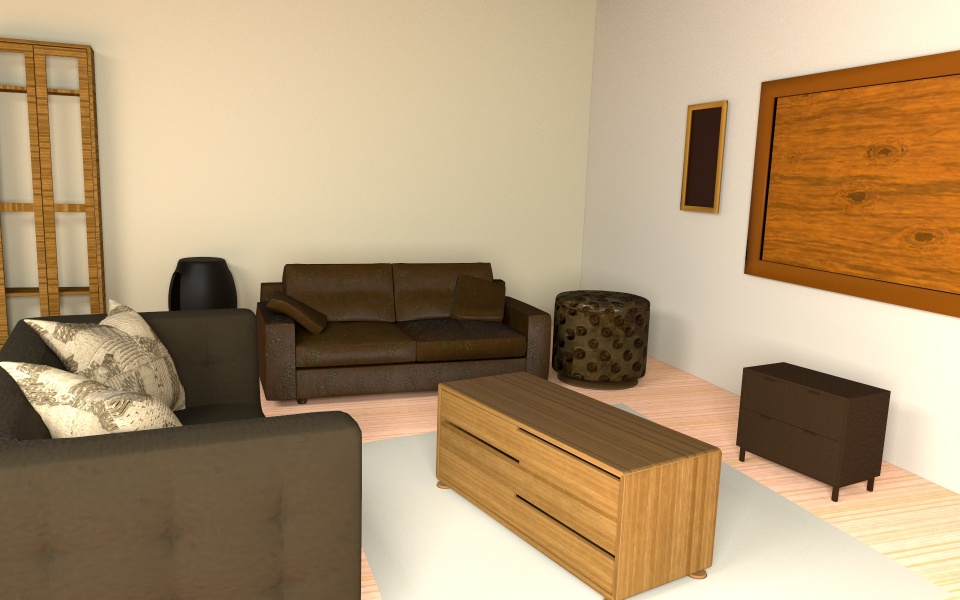} &
        \includegraphics[width=\resLen]{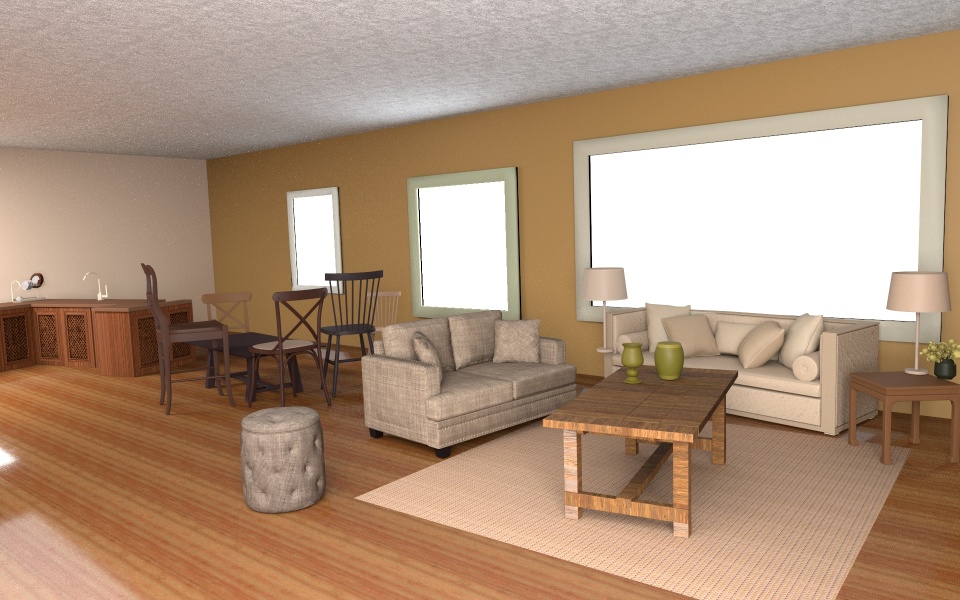}
        \\[-2pt]
        \emph{LPIPS \& RMSE:} \textbf{0.584} / \textbf{0.037} & \textbf{0.658} / \textbf{0.038} & \textbf{0.611} / \textbf{0.039} & \textbf{0.613} / \textbf{0.039} & \textbf{0.687} / \textbf{0.039}
        \\[2pt]
        \begin{overpic}[width=\resLen]{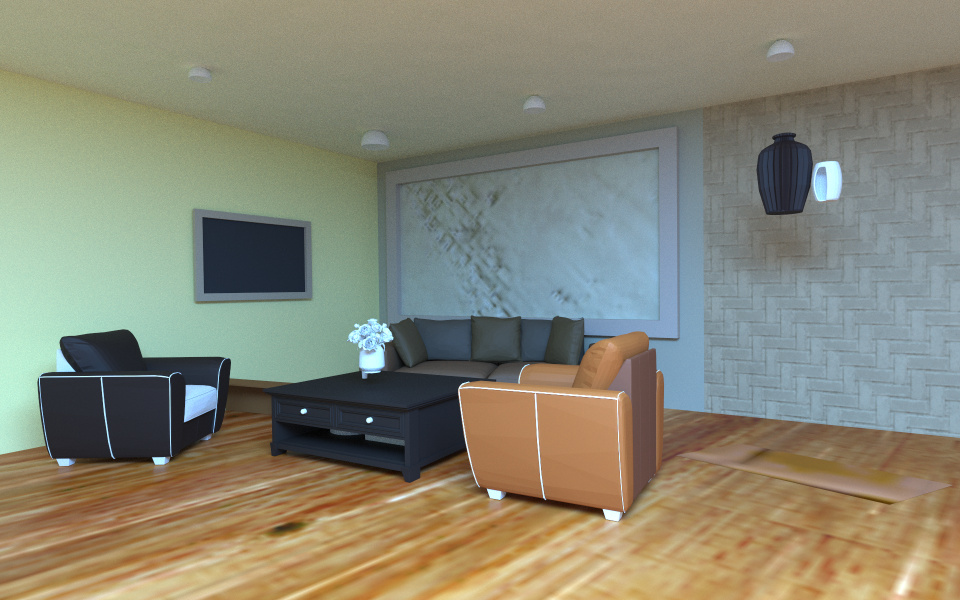}
            \put(2, 4){\color{white} \contour{black}{(c) \textbf{Ours} + \textsf{Photoscene}}}
        \end{overpic}
        &
        \includegraphics[width=\resLen]{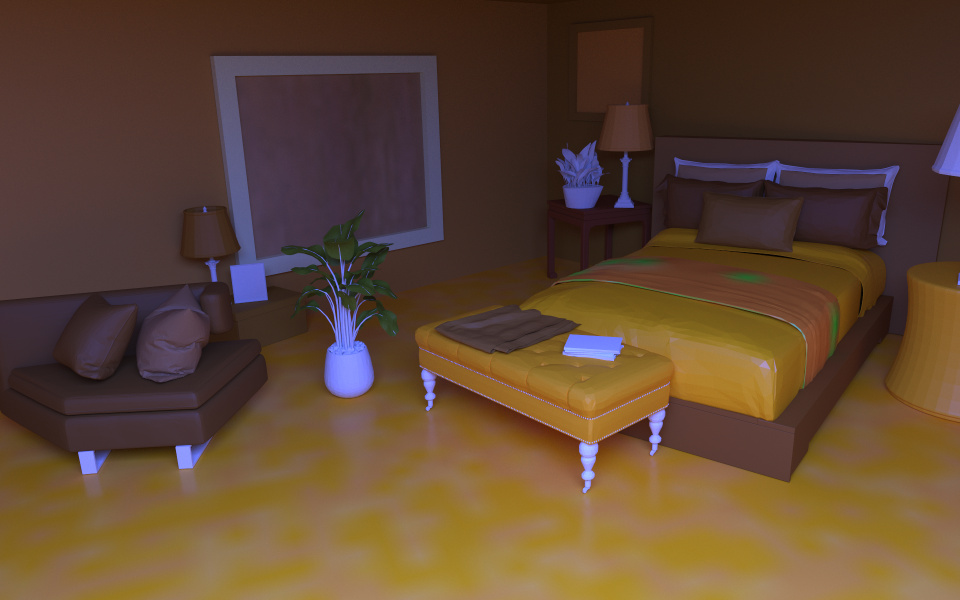} &
        \includegraphics[width=\resLen]{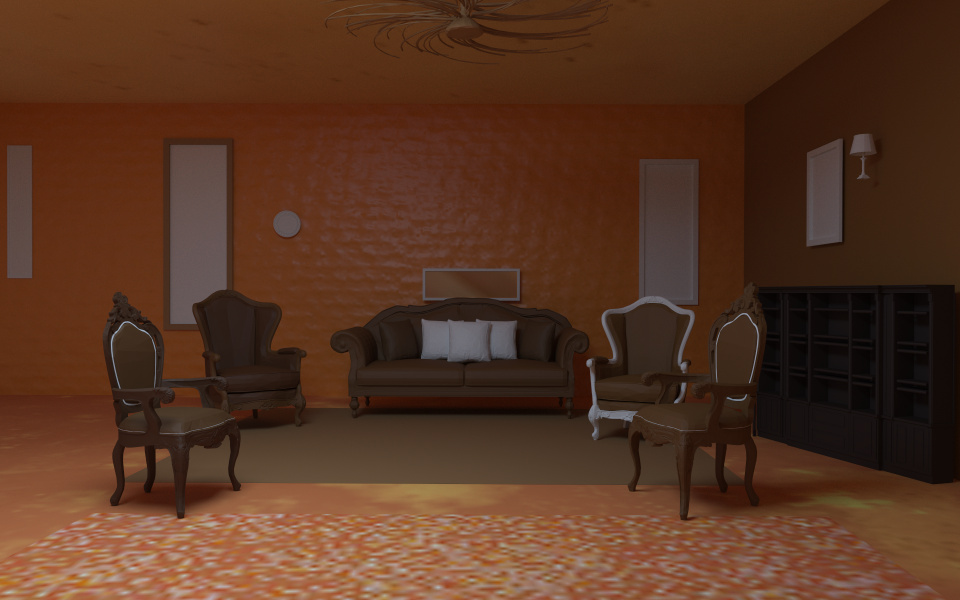} &
	\includegraphics[width=\resLen]{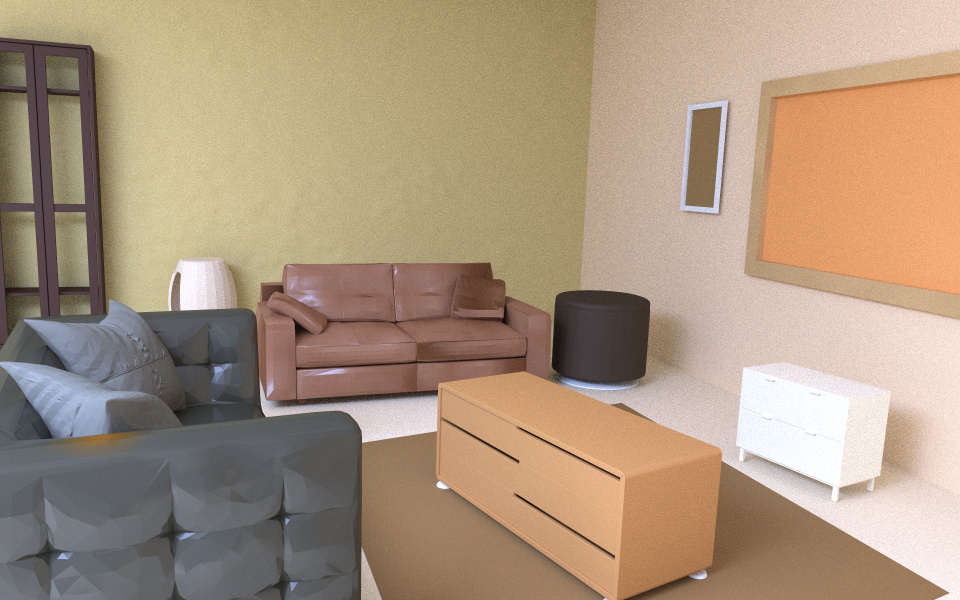} &
 	\includegraphics[width=\resLen]{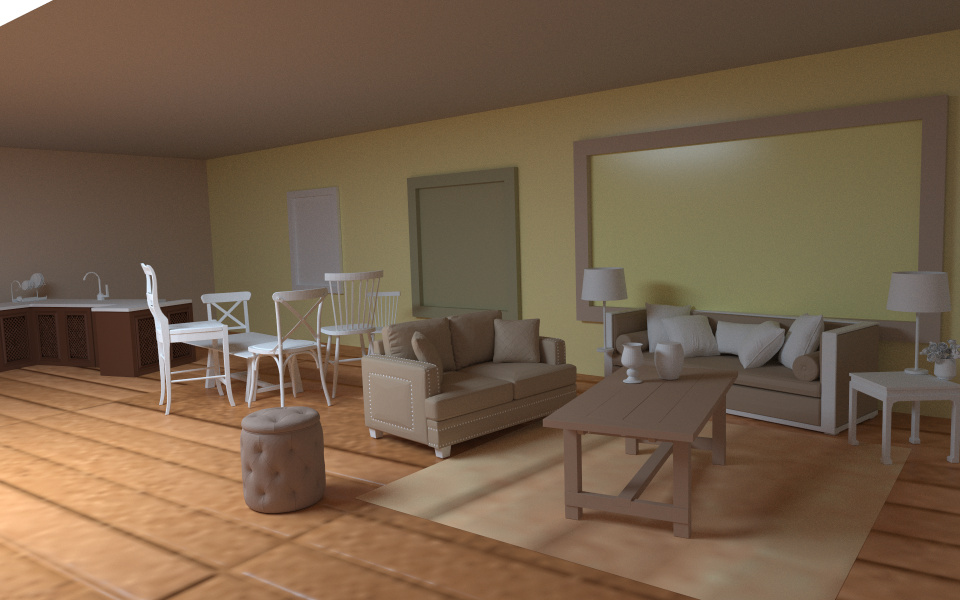}
        \\[-2pt]
        \emph{LPIPS \& RMSE:} 0.712 / 0.041 & 0.819 / 0.041 & 0.734 / 0.040 & 0.877 / 0.040 & 0.735 / \textbf{0.039}
        \\[2pt]
	\begin{overpic}[width=\resLen]{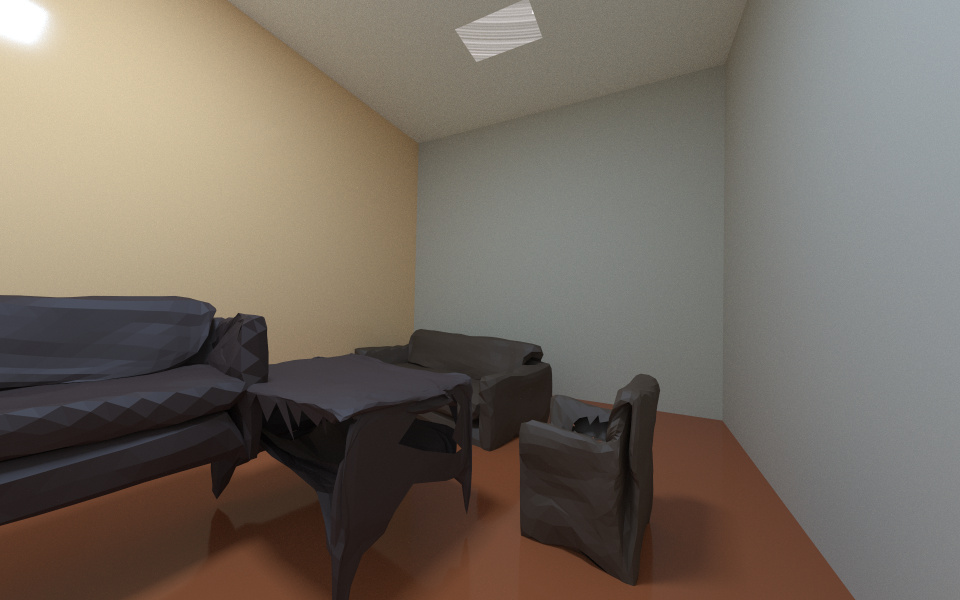}
            \put(2, 4){\color{white} \contour{black}{(d) \textsf{Total3D} + \textsf{Photoscene}}}
        \end{overpic}
        &
        \includegraphics[width=\resLen]{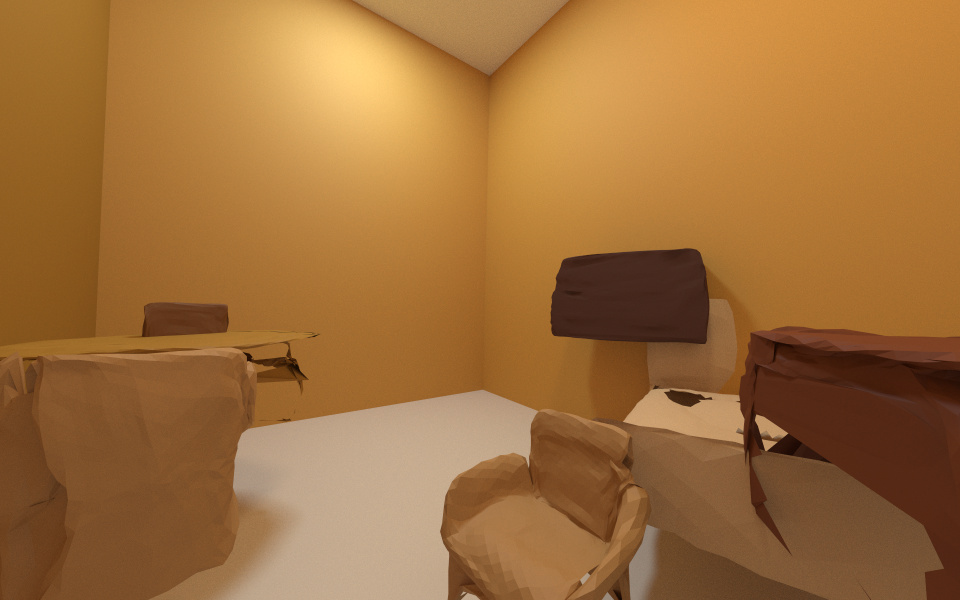} &
        \includegraphics[width=\resLen]{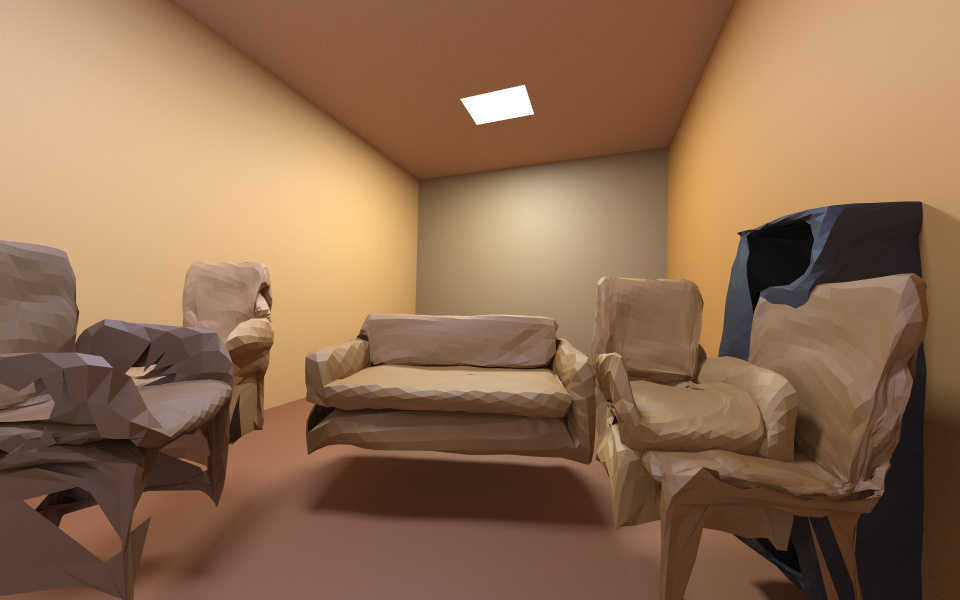} &
        \includegraphics[width=\resLen]{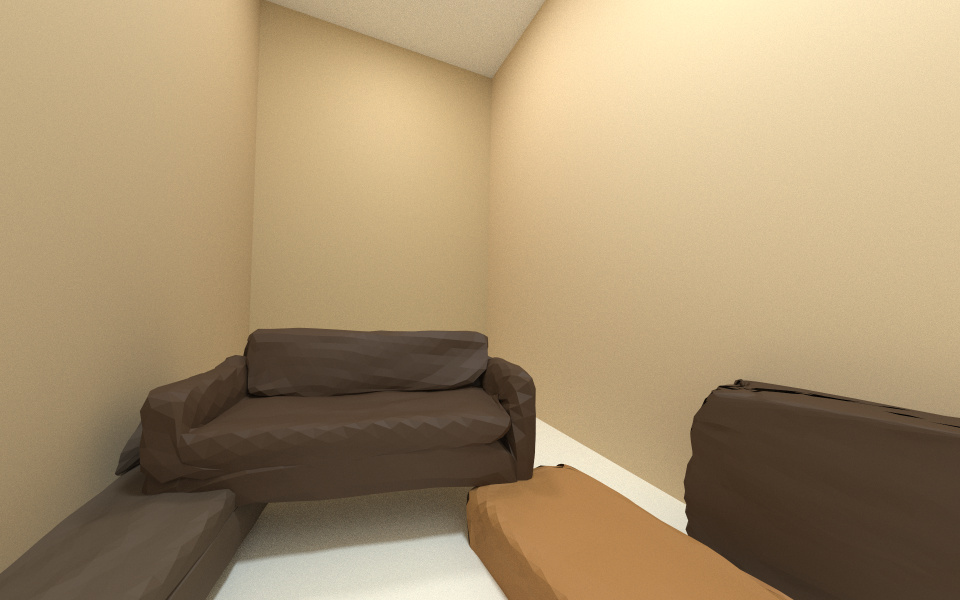} &
        \includegraphics[width=\resLen]{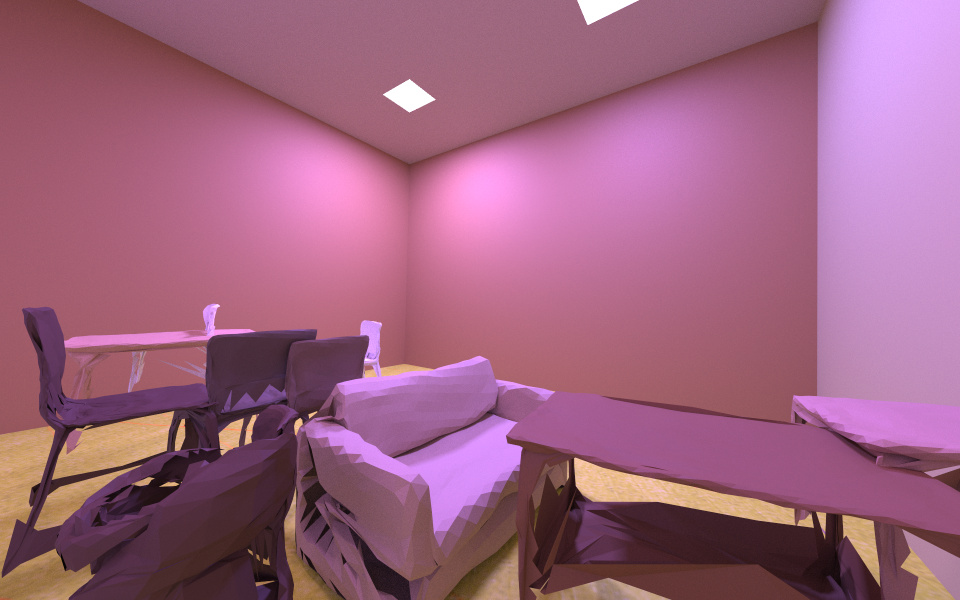}
        \\[-2pt]
        \emph{LPIPS \& RMSE:} 0.800 / 0.040 & 0.768 / 0.040 & 0.781 / 0.040 & 0.822 / 0.040 & 0.809 / 0.040
    \end{tabular}
    \caption{\label{fig:results}\rev{
        \textbf{Results and comparisons} for five clean input photos of room scenes (a). We compare our (automated) pipeline (b), our geometric pipeline combined with \textsf{PhotoScene}'s material and lighting pipeline (c), and \textsf{Total3D}'s geometry prediction combined with \textsf{PhotoScene}'s material and lighting (d).
        LPIPS and RMSE errors are shown under the images, with the lowest error shown in bold.
        Our pipeline (b) offers the best visual and numeric quality for all examples.
        \textsf{PhotoScene} with our geometry (c) produces numerically good results, but visually worse than ours, due to factors we discuss in the main text.
        \textsf{Total3D} produces lowest-quality geometry (d), though it should be noted that it is solving a much harder problem of neurally predicting the meshes, rather than searching for them in a database.
    }}
\end{figure*}

\rev{
    \paragraph*{Comparison with \textsf{IM2CAD}}
    We compare to \textsf{IM2CAD}~\cite{im2cad} in Fig.~\ref{fig:compare_im2cad}.
    Since the implementation of this technique has not been made publicly available, we apply our pipeline to two examples from their paper.
    To evaluate the results quantitatively, we use the LPIPS~\cite{zhang2018perceptual} metric, which is appropriate for comparing images with some misalignment, and RMSE (computed using downsampled images) metrics.
    The quantitative errors are shown under the images, with the lowest error shown in bold, showing that our approach better matches the target photographs.}
\rev{
    \paragraph*{Comparison with \textsf{PhotoScene}}
    In Fig.~\ref{fig:results}, we compare results generated with our technique (b) and two pipelines (c, d) based on \textsf{PhotoScene}~\cite{photoscene}.
    Similar to the comparison with \textsf{IM2CAD} in Fig.~\ref{fig:compare_im2cad}, we compare LPIPS and RMSE metrics to show the effectiveness of our method.
    \textsf{PhotoScene} with our geometry (c) produces numerically good results, but worse than ours, due to several factors. They used simpler approximations for differentiable rendering compared to our path-space system. We also found that comparing texture on rectangular crops is more robust than the masked VGG losses used by \textsf{PhotoScene}. Further, our CLIP based search provides better results than VGG-based Gram matrix distance as demonstrated on flat surfaces in the supplemental materials.
    When using \textsf{PhotoScene} with geometries reconstructed by \textsf{Total3D} (d), the reconstruction quality becomes poor, although it should be noted that \textsf{Total3D}~\cite{total3D} is solving a much harder problem of directly predicting the meshes, rather than searching for them in a database.
}

\begin{figure*}[t]
    \centering
    \small
    \setlength{\resLen}{1.75in}
    \addtolength{\tabcolsep}{-4pt}
    \begin{tabular}{cccc}
        (a1) \textbf{Target} (w/ user crops) & (b1) \textbf{Ours} (w/ user crops) &
        (a2) \textbf{Target} (w/ auto crops) & (b2) \textbf{Ours} (w/ auto crops)
        \\
        \includegraphics[trim=0 0 0 200,clip,width=\resLen]{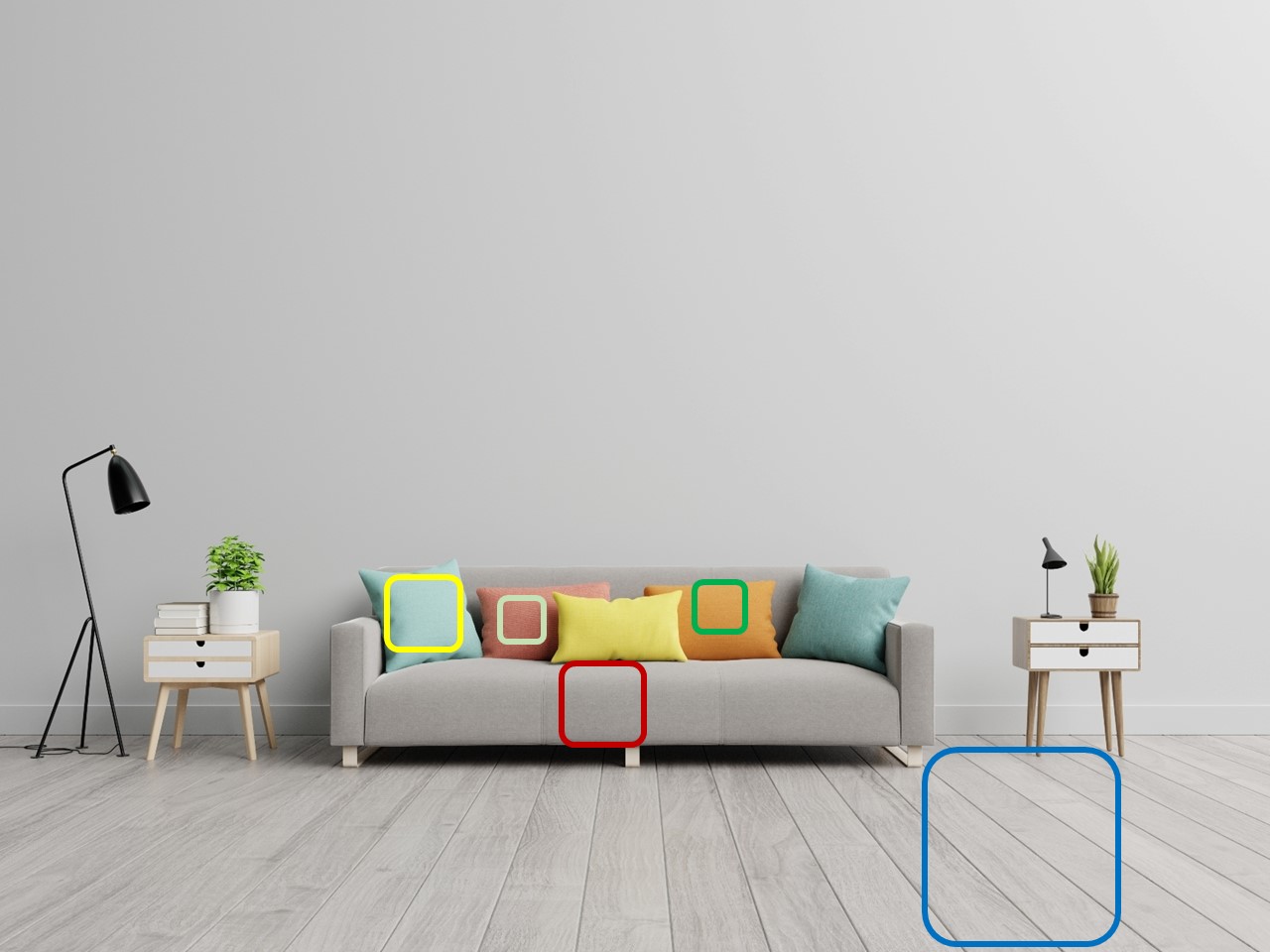} &
        \includegraphics[trim=0 0 0 200,clip,width=\resLen]{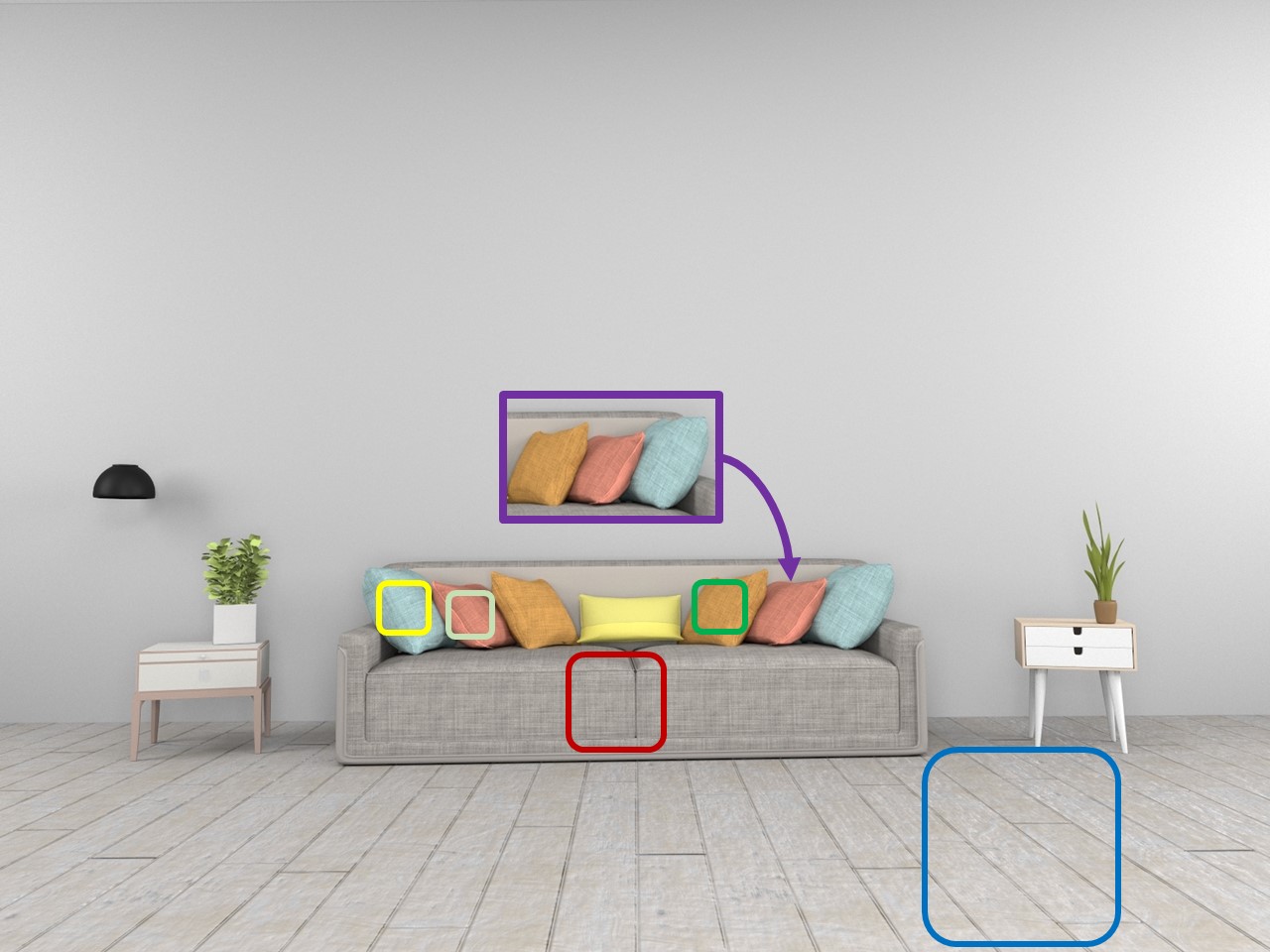} &
        \includegraphics[trim=0 0 0 200,clip,width=\resLen]{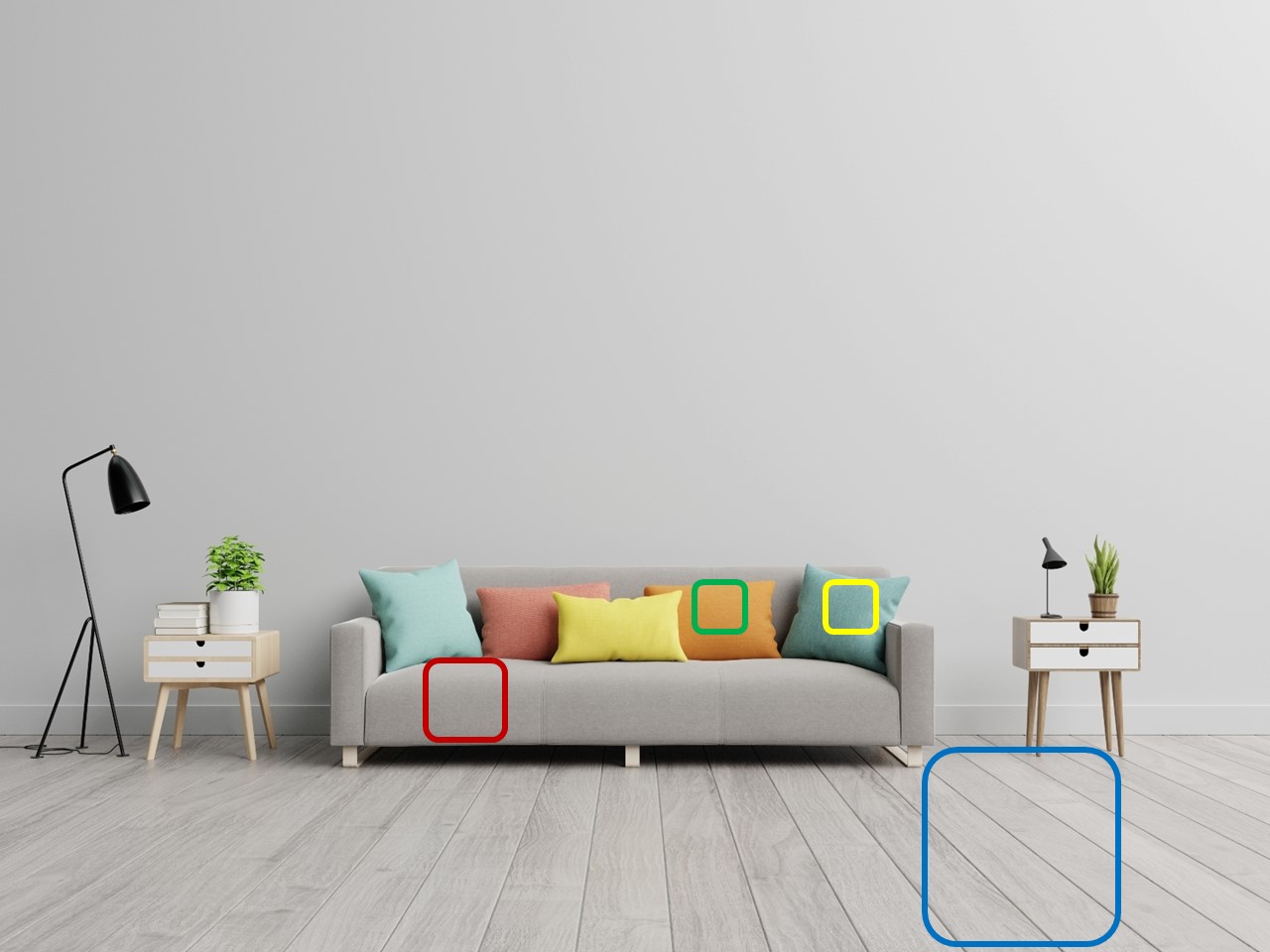} &
        \includegraphics[trim=0 0 0 200,clip,width=\resLen]{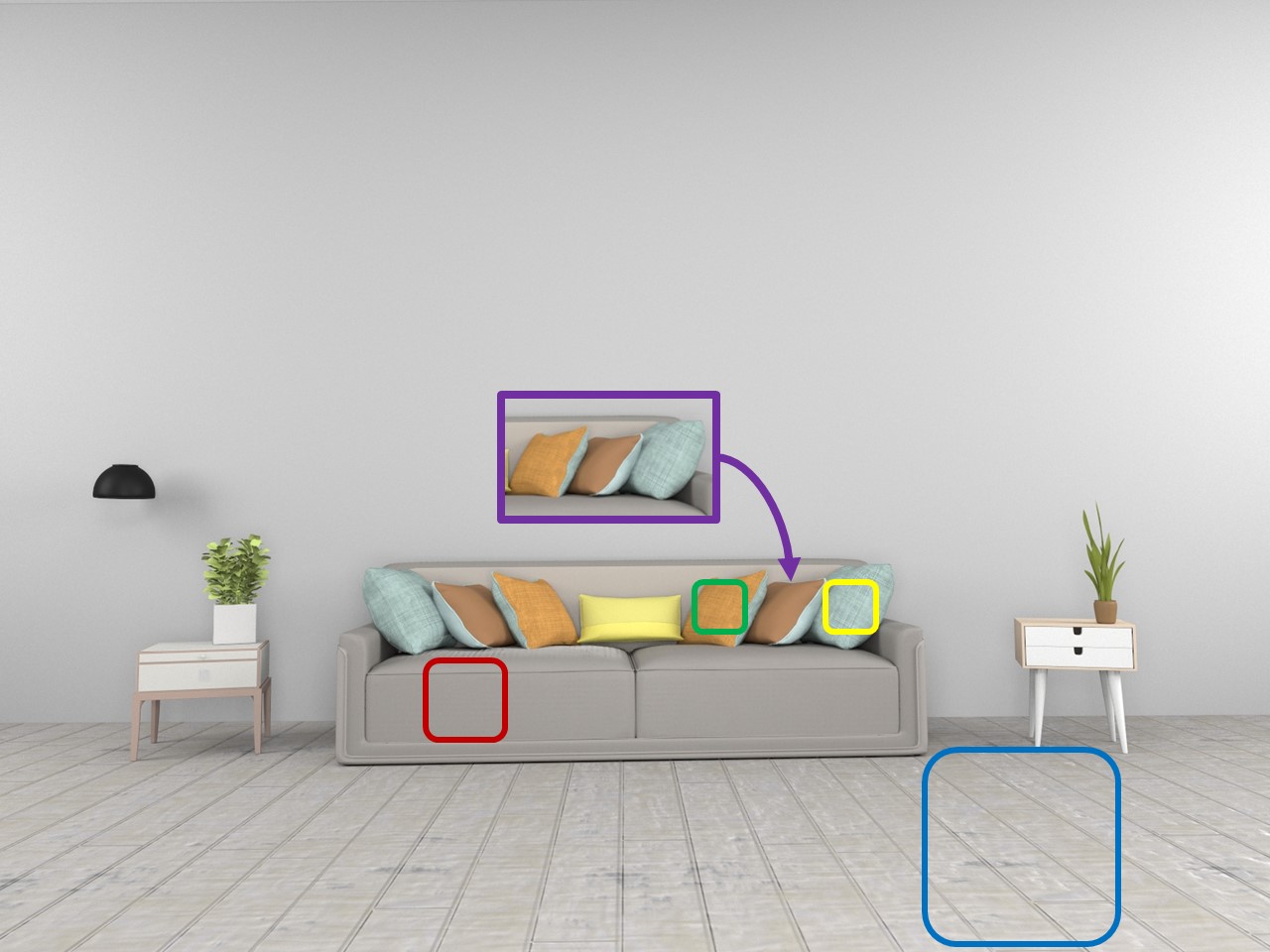} 
        \\
        \includegraphics[width=\resLen]{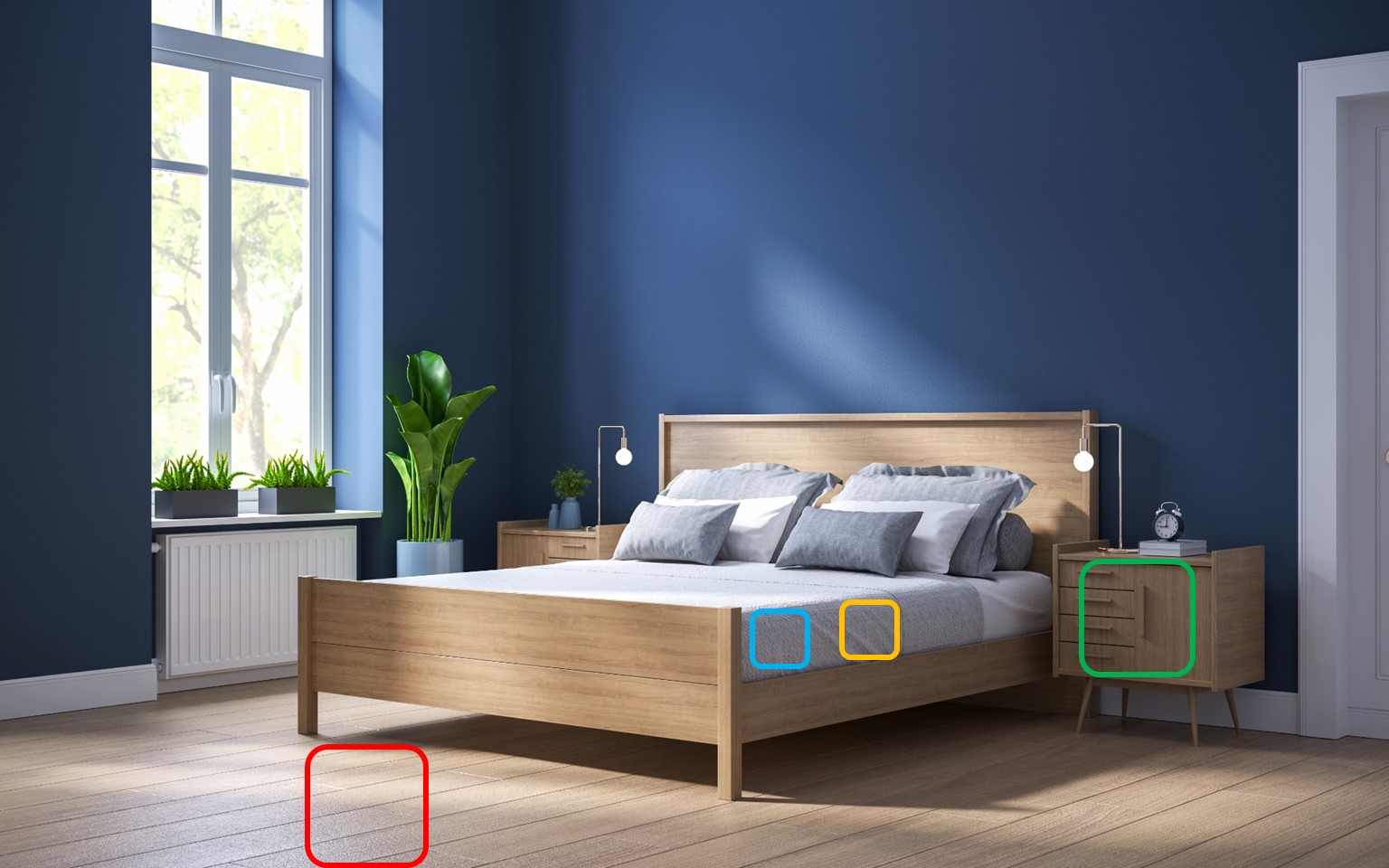} &
        \includegraphics[width=\resLen]{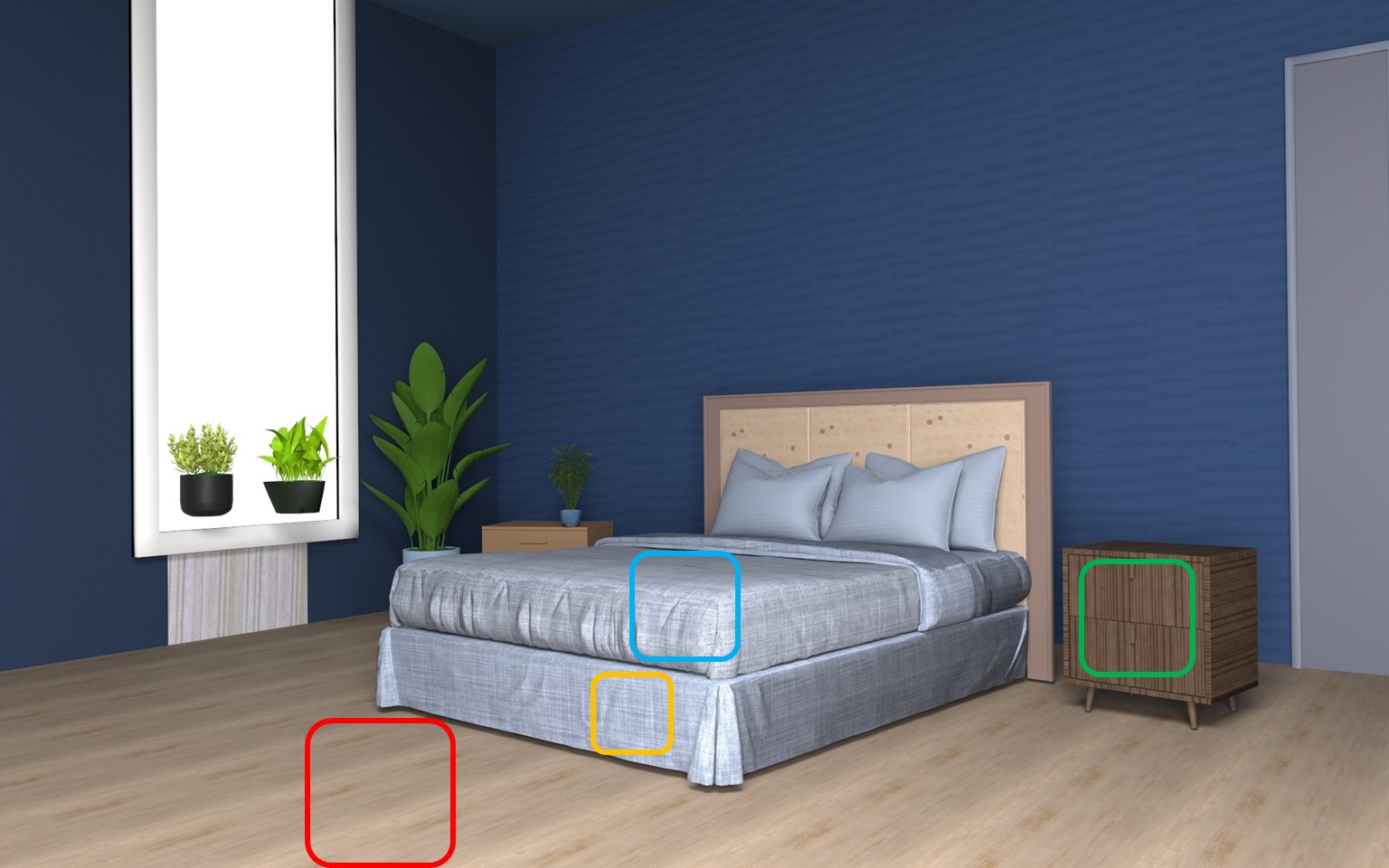} &
        \includegraphics[width=\resLen]{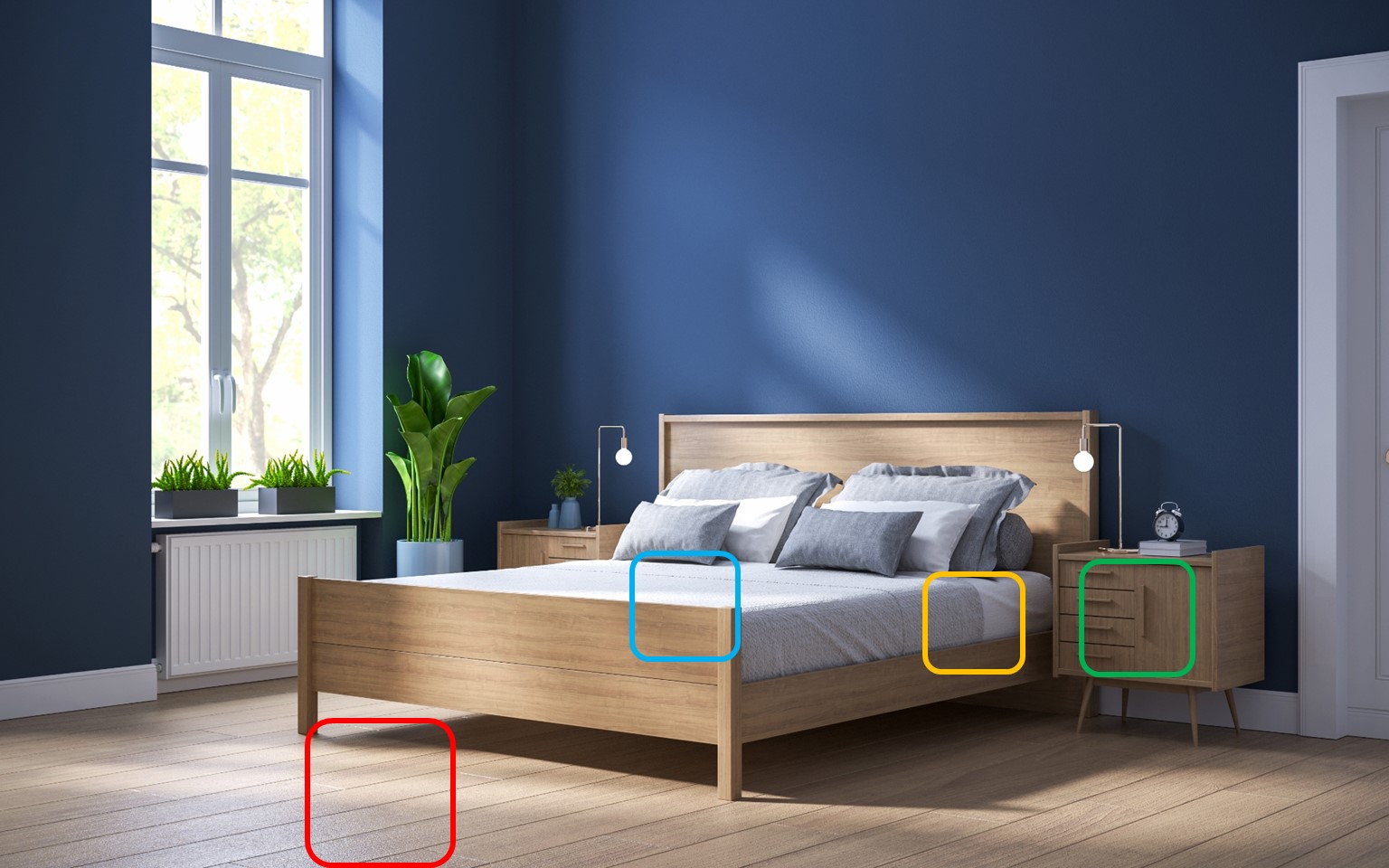} &
        \includegraphics[width=\resLen]{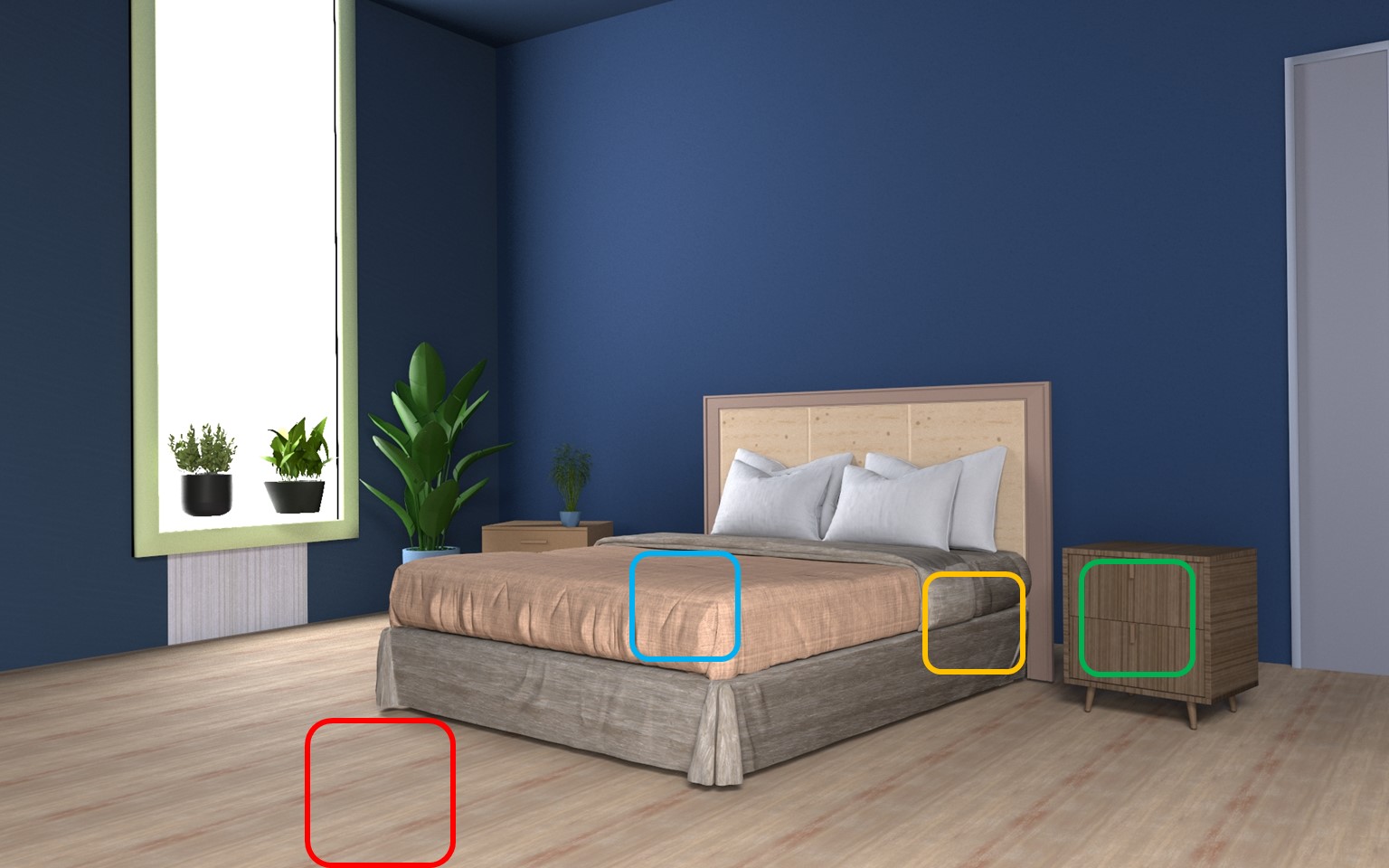}
    \end{tabular}
    \caption{\label{fig:user_crop}\rev{%
    In the case where some correspondences are not trivial to establish (i.e between the different pillows (top row)) or the user desires to match a geometry to a different material in the scene (bottom row), user-specified crops can be used, potentially better matching their intent than our automatic crop selection.}}
\end{figure*}

\rev{%
    \paragraph*{Using user-specified crops}
    Results generated by our automated pipeline can be further improved by using minimal user inputs.
    We demonstrate this in Fig.~\ref{fig:user_crop} where user-specified crops are used to drive the material and lighting stage (\S\ref{ssec:finalstage}).
}

\paragraph*{Performance and supplementary materials}
On average, the room optimization stage takes 100 iterations, with less than 0.1s per iteration. The geometry initialization and CLIP search take usually around 5s per object in the scene. \revky{The optimization time per iteration is around 1-3 seconds, with a total of 300 iterations. The fine stage usually take 3-5 seconds per iteration, with about 120 iterations.} A more detailed performance table is shown in the supplementary materials. The supplementary also shows animations illustrating optimization progress and partial results.

\section{Limitations and Future Work}
\revky{While PSDR-Room already gives useful results, there are some challenges left.
First, while our fully automatic pipeline outperforms previous work, it is not perfect, and can sometimes benefit from user inputs. For example, user-provided crop pairs can focus attention on matching materials that matter to a human viewer, and establish correspondences between material parts that are semantically related but not aligned in image space. Further, a user picking from, say, top-3 materials and objects returned by CLIP search can sometimes make a better choice than simply using the highest-ranked result.

Our current box assumption is simple, and could be extended to more intricate floor plan layouts, as well as non-horizontal ceilings. Our work pushes even the most advanced differentiable rendering approaches to their limits. Intersection between objects causes erroneous zero gradient due to a missing boundary term. Differentiable collision detection during the optimization process could help avoid two objects intersecting, but remains out of the scope of this work. 

Object databases with well-separated material parts are not easy to come by and our material database is fairly small; a commercial deployment of our method would likely invest in curating larger databases of objects and materials with the right properties.}

\section{Conclusion}
\label{sec:conclusion}
In this paper, we introduced \sysName, an end-to-end pipeline that generates 3D models of indoor scenes, complete with objects, lighting and textured materials, with minimal user input.
Based on a single segmented image, our technique uses CLIP search to automatically select the shapes and materials for each object from databases.
Further, our pipeline refines the object poses, lighting  and materials of each object by leveraging physics-based differentiable rendering, significantly improving the reconstruction quality compared to previous work when evaluated on real indoor photographs.
Our generated 3D models can be easily edited by the user as a post-process. We believe our pipeline provides a solid foundation for future improvement towards better learning-based initialization approaches, more complex room layouts, or other environments beyond room scenes.

\bibliographystyle{ACM-Reference-Format}
\bibliography{PSDR_arxiv}
\clearpage
\begin{figure}[t]
    \centering
    \includegraphics[width=\columnwidth]{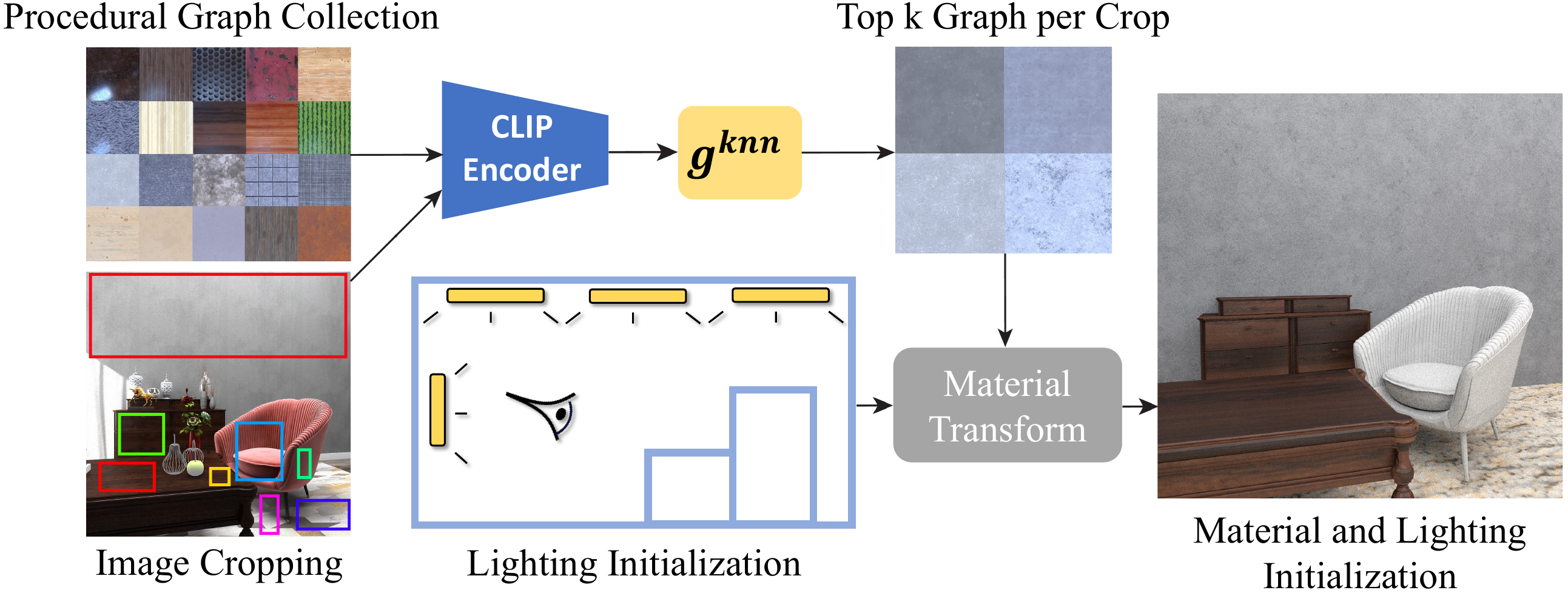}
    \caption{\label{fig:Mat_Init}
        We use CLIP for zero-shot classification and ranking of materials, by comparing encoded thumbnails of our procedural materials with encoded input image crops. A user can optionally pick from top $k$ matches. We further search for the best transform (scale and rotation) for the best texture match; this is done under the initial lighting. We keep some materials homogeneous (more detail in text).
    }
\end{figure}

\begin{figure}[t]
    \centering
    \includegraphics[width=\columnwidth]{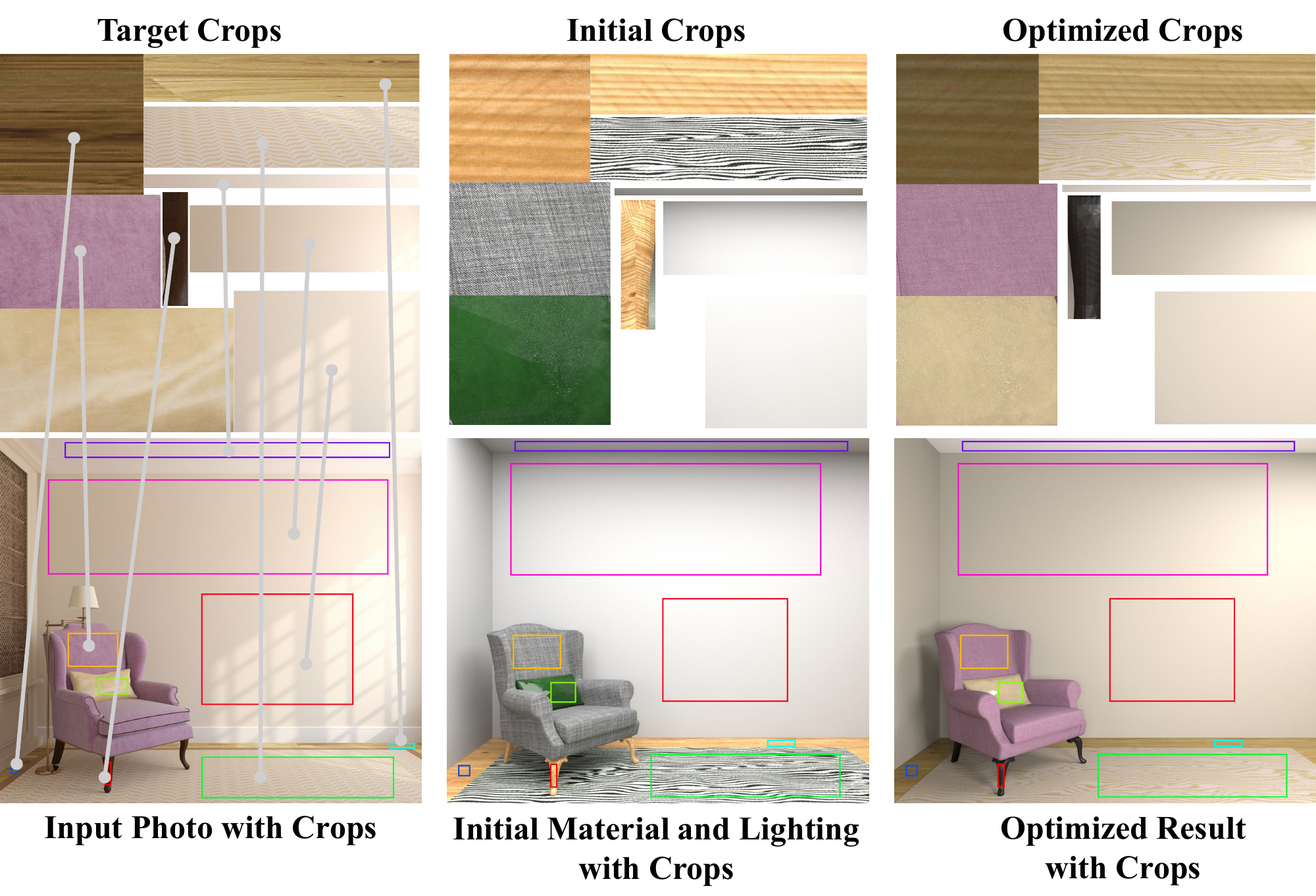}
    \caption{\label{fig:crop_result}
        We visualize the crops from the input image, paired with rendered crops at initialization and during optimization. The optimized crops match the mean color and texture better comparing to the initialization. Losses based on Gram matrices of VGG layers are effective at matching texture in rectangular crops without need for pixel alignment.
    }
\end{figure}



\end{document}